\documentclass[final,3p,times,twocolumn]{elsarticle}

\usepackage{amssymb}
\usepackage{amsmath}
\usepackage{amsthm}

\usepackage{graphicx}
\graphicspath{{figures/}}

\usepackage{booktabs}
\usepackage{multirow}

\usepackage{hyperref}
\hypersetup{
    colorlinks=true,
    linkcolor=blue,
    citecolor=blue,
    urlcolor=blue
}

\usepackage{lineno}

\usepackage{subcaption}

\usepackage[table,svgnames]{xcolor}

\definecolor{cellgreen1}{rgb}{0.50,0.90,0.60}  
\definecolor{cellgreen2}{rgb}{0.65,0.94,0.70}  
\definecolor{cellgreen3}{rgb}{0.78,0.97,0.82}  
\definecolor{cellyellow}{rgb}{0.99,0.96,0.76}  
\definecolor{cellorange}{rgb}{0.98,0.86,0.76}  

\makeatletter
\long\def\printFirstPageNotes{%
  \iflongmktitle
    \let\columnwidth=\textwidth
  \fi
\ifdoubleblind
\else
  \ifx\@tnotes\@empty\else\@tnotes\fi
  \ifx\@nonumnotes\@empty\else\@nonumnotes\fi
  \ifx\@cornotes\@empty\else\@cornotes\fi
  \ifx\@elseads\@empty\relax\else
   \let\thefootnote\relax
   \footnotetext{\hangindent1.8em\hangafter1
     \ifnum\theead=1\relax
      \textit{Email address:\space}\else
      \textit{Email addresses:\space}\fi
     \@elseads}\fi
  \ifx\@elsuads\@empty\relax\else
   \let\thefootnote\relax
   \footnotetext{\hangindent1.8em\hangafter1
     \textit{URL:\space}%
     \@elsuads}\fi
\fi
  \ifx\@fnotes\@empty\else\@fnotes\fi
  \iflongmktitle\if@twocolumn
   \let\columnwidth=\Columnwidth\fi\fi
}
\makeatother

\makeatletter
\long\def\MaketitleBox{%
  \resetTitleCounters
  \def\baselinestretch{1}%
  \begin{\elsarticletitlealign}%
   \def\baselinestretch{1}%
    \Large\@title\par\vskip10pt
  \ifdoubleblind
    \vspace*{2pc}
  \else
    \normalsize\elsauthors\par\vskip8pt
    \footnotesize\itshape\elsaddress\par\vskip8pt
  \fi
    \vskip2pt
    \includegraphics[width=0.92\textwidth,height=0.22\textheight,keepaspectratio]{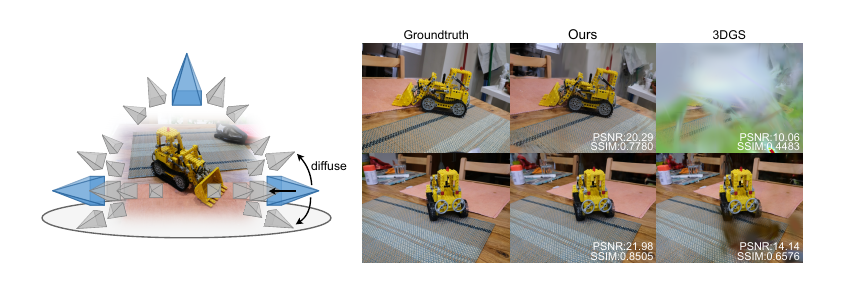}\par
    \refstepcounter{figure}\label{fig:teaser}%
    \vskip2pt
    \parbox{\textwidth}{\footnotesize\textbf{Figure~\thefigure:} Overview of our diffusion-assisted sparse-view 3DGS framework. \emph{Left:} given only a few input views (red), we score and select extrapolated pseudo-view cameras (blue) through Generative Active Pseudo-view Selection, and synthesize geometrically consistent images with a constrained image diffusion prior. \emph{Right:} on representative held-out test views, our reconstructions retain sharp object boundaries and clean backgrounds where vanilla 3DGS~\cite{kerbl20233dgs} collapses into floaters and washed-out patches, with consistent PSNR and SSIM gains under the same sparse input set.}\par
    \vskip6pt
    \hrule\vskip6pt
    \ifvoid\absbox\else\unvbox\absbox\par\vskip6pt\fi
    \ifvoid\keybox\else\unvbox\keybox\par\vskip6pt\fi
    \hrule\vskip6pt
  \end{\elsarticletitlealign}%
}
\makeatother

\begin{document}

\begin{frontmatter}

\title{GAPS: Generative Active Pseudo-view Selection for Sparse-View 3D Gaussian Splatting}


\author[inst1,inst2]{Hongfei Zhu\corref{cor1}}
\cortext[cor1]{Corresponding author}
\ead{vit_oo@sjtu.edu.cn}

\author[inst3,inst2]{Haochen Deng}
\ead{24210720101@m.fudan.edu.cn}

\author[inst2]{Sitao Zhang}
\ead{alex.zhangsit@goertek.com}

\author[inst2]{Ling Zhou}
\ead{ling.zhou@goertek.com}

\affiliation[inst1]{
    organization={Shanghai Jiao Tong University},
    addressline={800 Dongchuan Road},
    city={Shanghai},
    postcode={200240},
    state={Shanghai},
    country={China}
}

\affiliation[inst2]{
    organization={Goertek Inc.},
    addressline={No.~268 Dongfang Road, Hi-tech Industrial Development Zone},
    city={Weifang},
    postcode={261031},
    state={Shandong},
    country={China}
}

\affiliation[inst3]{
    organization={Fudan University},
    addressline={220 Handan Road},
    city={Shanghai},
    postcode={200433},
    state={Shanghai},
    country={China}
}
\begin{abstract}
Novel view synthesis from sparse observations remains a significant challenge due to the severely under-constrained nature of the problem. 3D Gaussian Splatting (3DGS), despite its efficiency in real-time rendering, produces severe artifacts---floaters, broken geometry, and washed-out backgrounds---when only a few input views are available. We propose an alternating optimization framework that leverages pre-trained image diffusion models to generate geometrically consistent pseudo-views as additional supervision for 3DGS training. The generative process is constrained through depth-conditioned ControlNet, IP-Adapter style transfer, LoRA scene adaptation, and img2img structural anchoring. To determine \emph{where} to generate pseudo-views, we introduce Generative Active Pseudo-view Selection (GAPS), a scoring function that jointly balances reconstruction informativeness and generative reliability. GAPS employs an annealing schedule that transitions from conservative interpolation in early training---where generation quality is prioritized---to exploratory extrapolation in later stages, progressively expanding the coverage of unobserved regions. A dual-criterion admission gate with uncertainty-weighted losses further prevents unreliable generations from destabilizing training, while density-adaptive DropGaussian regularization mitigates overfitting in geometrically complex scenes. Experiments on LLFF (3/6/9 views) and Mip-NeRF~360 (12/24 views) demonstrate consistent improvements over vanilla 3DGS, with average PSNR gains of $+0.40$/$+0.89$/$+0.70$~dB on LLFF and $+1.18$/$+0.80$~dB on Mip-NeRF~360, accompanied by uniform SSIM increases and LPIPS reductions across all evaluated settings. Component ablation confirms that the scored active selection and the density-adaptive regularizer are individually necessary, and that only the full method improves LPIPS below the no-pseudo-view baseline on unbounded $360^{\circ}$ scenes.
\end{abstract}




\end{frontmatter}



\section{Introduction}
\label{sec:introduction}
Figure~\ref{fig:teaser} provides a visual overview of the proposed GAPS framework and its qualitative effect under sparse-view supervision.
Novel view synthesis (NVS) is a fundamental problem in computer vision with broad applications in virtual reality, autonomous driving, and digital content creation. 3D Gaussian Splatting (3DGS)~\cite{kerbl20233dgs} has emerged as a leading approach, representing scenes with explicit Gaussian primitives rendered via differentiable rasterization at real-time frame rates. However, 3DGS relies on dense multi-view observations; when only a sparse set of input views is available, the reconstruction becomes severely under-constrained, producing artifacts such as erroneous depth, floating structures, and inconsistent appearance. The adaptive densification of 3DGS further exacerbates this issue, as unobserved regions receive no gradient signal and remain unrecovered.

Existing sparse-view methods either regularize the optimization process~\cite{li2024dngaussian, zhang2024cor_gs, park2025dropgaussian} or leverage pre-trained diffusion models as generative priors~\cite{kong2025gsgs, topaloglu2025oraclegs}. Regularization-based approaches mitigate overfitting but cannot synthesize content for unobserved regions. Diffusion-assisted methods, notably GS-GS~\cite{kong2025gsgs}, generate pseudo-views as additional training supervision through alternating optimization with a scene-adapted diffusion model. However, key questions remain: \emph{where} to generate pseudo-views for maximum benefit, \emph{how} to prevent low-quality generations from destabilizing training, and \emph{how} to adapt regularization to scene-specific complexity.

We propose an alternating optimization framework that addresses these gaps. Our method constrains the diffusion generation through four complementary mechanisms: ControlNet~\cite{zhang2023controlnet} for depth-conditioned geometric structure, IP-Adapter~\cite{ye2023ipadapter} for stylistic coherence, LoRA~\cite{hu2022lora} fine-tuning for scene-specific appearance, and img2img denoising to anchor the coarse structure of Gaussian renderings. To determine the most beneficial viewpoints, we introduce \emph{Generative Active Pseudo-view Selection} (GAPS), a scoring function that jointly balances reconstruction informativeness and generative reliability with an annealing schedule that progresses from conservative interpolation to exploratory extrapolation. We further propose \emph{density-adaptive DropGaussian} regularization that scales dropout strength according to scene complexity, and a \emph{dual-criterion admission} mechanism with \emph{uncertainty-weighted losses} that gates and down-weights unreliable pseudo-views.

The main contributions are:
\begin{itemize}
    \item Generative Active Pseudo-view Selection (GAPS) that jointly optimizes for reconstruction informativeness and generative reliability with progressive extrapolation tiers.
    \item An alternating GS-diffusion optimization framework with four-fold constrained pseudo-view generation (ControlNet, IP-Adapter, LoRA, img2img) that restricts diffusion outputs to geometrically and stylistically consistent images.
    \item Density-adaptive DropGaussian regularization and dual-criterion pseudo-view admission with uncertainty-weighted losses for robust training.
\end{itemize}

\section{Related Work}
\label{sec:related}

\subsection{Sparse-View 3D Reconstruction}
\label{sec:related_sparse}

Neural Radiance Fields (NeRF)~\cite{mildenhall2020nerf} and its extensions~\cite{barron2022mipnerf360} achieve photorealistic novel view synthesis through continuous volumetric representations but require dense input views and slow per-ray sampling. 3D Gaussian Splatting (3DGS)~\cite{kerbl20233dgs} replaces volumetric rendering with explicit Gaussian primitives and differentiable rasterization, enabling real-time rendering. However, its adaptive densification relies on view-space gradients from observed cameras, leaving unobserved regions without reconstruction signal.

Under sparse-view conditions, both paradigms degrade significantly. NeRF-based methods introduce regularization through semantic consistency~\cite{jain2021dietnerf}, patch-based appearance priors~\cite{niemeyer2022regnerf}, frequency masking~\cite{yang2023freenerf}, and depth ranking distillation~\cite{wang2023sparsenerf}. In the 3DGS domain, DNGaussian~\cite{li2024dngaussian} regularizes Gaussian depth with monocular estimates, FSGS~\cite{zhu2024fsgs} densifies under-represented regions via proximity-guided unpooling, CoR-GS~\cite{zhang2024cor_gs} enforces multi-view consistency through co-regularization, and DropGaussian~\cite{park2025dropgaussian} applies random primitive dropout to reduce overfitting. While effective, these methods operate solely within the information available from observed viewpoints and cannot synthesize content for unobserved regions.

\subsection{Diffusion-Assisted 3D Reconstruction}
\label{sec:related_diffusion}

Pre-trained image diffusion models~\cite{rombach2022ldm, podell2024sdxl} encode rich priors about natural image statistics, making them promising for compensating sparse-view information deficits. DreamFusion~\cite{poole2023dreamfusion} pioneered Score Distillation Sampling (SDS) to lift 2D diffusion priors into 3D, but SDS targets content generation rather than faithful reconstruction and produces over-smoothed results. More recent approaches directly generate pseudo-views as training supervision: OracleGS~\cite{topaloglu2025oraclegs} adopts a ``propose-and-validate'' paradigm where a generative model synthesizes novel views and an MVS model validates them via attention-derived uncertainty maps; Bose et al.~\cite{bose2025uar_scenes} refine single-image 3D reconstructions using entropy-based uncertainty from video diffusion models; Gaussian Scenes~\cite{paul2025gaussian_scenes} integrates depth-enhanced diffusion priors for pose-free reconstruction.

\textbf{GS-GS}~\cite{kong2025gsgs} is most closely related to our work. It introduces alternating optimization between 3DGS and a diffusion model: the Gaussian representation is optimized with both ground-truth and generated pseudo-views, while the diffusion model is adapted to the scene via LoRA fine-tuning~\cite{hu2022lora}. Pseudo-views are generated at interpolated poses using a depth T2I-adapter for geometric conditioning. A geometry-aware fine-tuning strategy warps pseudo-view renders to training views and enforces L1 consistency on diffusion features, and a depth regularization term aligns Gaussian-rasterized depth with monocular depth estimates via MS-SSIM. GS-GS achieves state-of-the-art results on the Blender, LLFF, and Mip-NeRF 360 datasets.

More broadly, diffusion-assisted sparse-view reconstruction methods---including GS-GS---share three open challenges. (1)~\emph{View selection}: pseudo-view cameras are typically placed at fixed interpolations between training pairs without considering which viewpoints would most benefit the reconstruction. As discussed in Section~\ref{sec:related_active}, existing active view selection methods~\cite{jiang2024fisherrf, li2024frequency_view} optimize purely for reconstruction gain, yet when the image source is a generative model rather than a real camera, the reliability of the generation must be jointly considered. (2)~\emph{Quality control}: generated pseudo-views often receive equal supervision weight with no mechanism to reject or down-weight geometrically inconsistent outputs. (3)~\emph{Scene adaptivity}: regularization strategies are typically uniform, despite scene complexity varying significantly (e.g., Gaussian counts ranging from 95k for an indoor room to 310k for dense foliage). Building on the alternating optimization paradigm of GS-GS, our work addresses these challenges: we introduce Generative Active Pseudo-view Selection (GAPS) that jointly balances informativeness and generative reliability---filling a gap in both sparse-view reconstruction and active view planning---dual-criterion admission with uncertainty-weighted losses for robust quality control, and density-adaptive DropGaussian regularization for scene-aware training.

\subsection{Active View Selection and Uncertainty Estimation}
\label{sec:related_active}

Active view selection seeks to determine the most informative camera placement for 3D reconstruction. FisherRF~\cite{jiang2024fisherrf} uses the Fisher information of radiance field parameters to quantify viewpoint informativeness, and Li et al.~\cite{li2024frequency_view} propose frequency-based criteria for Gaussian Splatting. However, these methods optimize purely for reconstruction gain without considering the reliability of the image source---a critical factor when views are generated by a diffusion model rather than captured by a real camera.

Uncertainty estimation in 3DGS has been explored through learned per-Gaussian uncertainty~\cite{kim2024uncertainty_4dgs}, predictive photometric confidence~\cite{galappaththige2026predictive_uncertainty}, and MVS attention-based validation~\cite{topaloglu2025oraclegs}. Our approach uses a lightweight residual-based uncertainty $u(\mathbf{p}) = |I_{\text{hall}}(\mathbf{p}) - I_{\text{GS}}(\mathbf{p})|$ that directly measures generation-reconstruction agreement without additional models, serving both as an admission gate and as per-pixel confidence weights during training.

\section{Methodology}
\label{sec:method}
\begin{figure*}[!t]
  \centering
  \includegraphics[width=\textwidth]{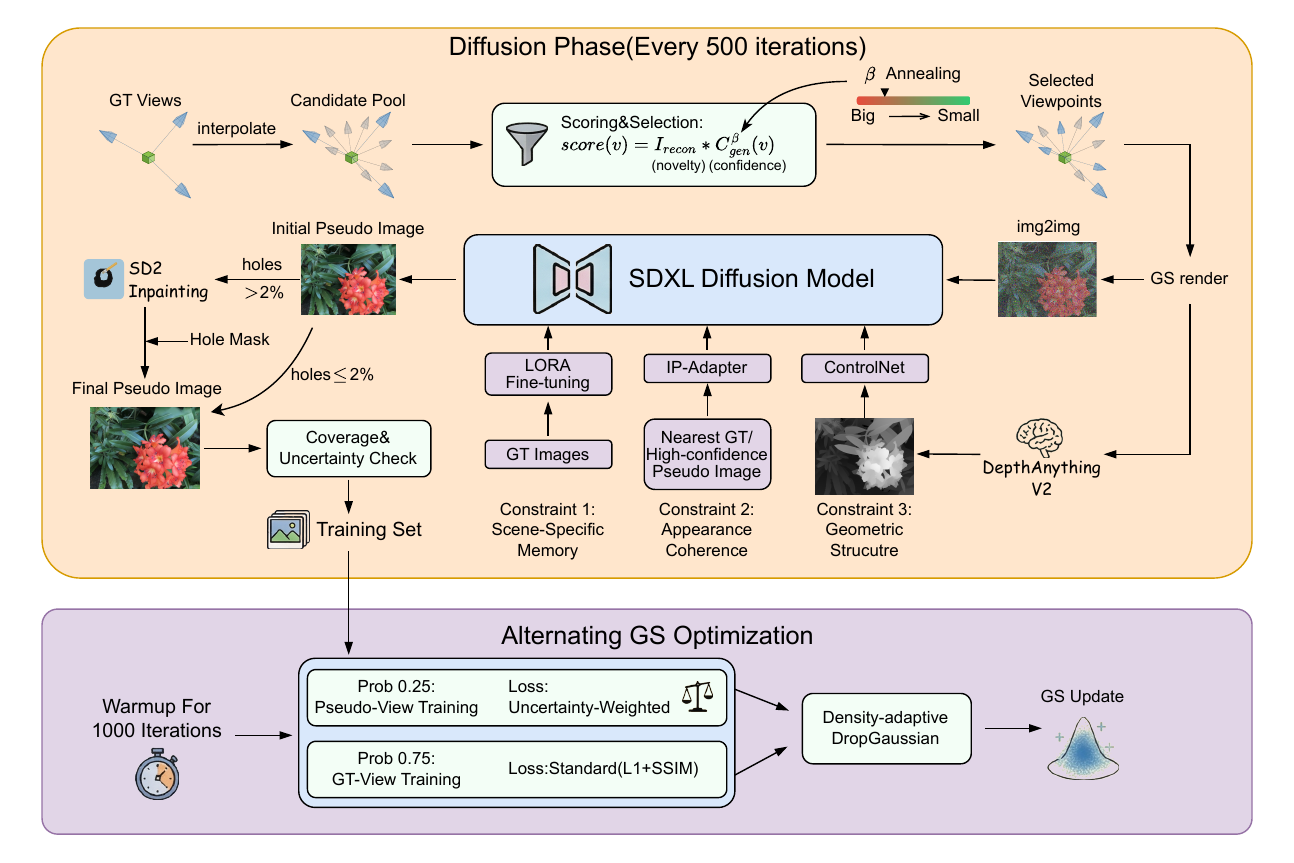}
  \caption{Overview of the proposed alternating optimization framework. Every $\Delta T\!=\!500$ Gaussian iterations, a \emph{diffusion phase} (top) selects extrapolated viewpoints from a candidate pool through Generative Active Pseudo-view Selection, scoring each candidate by reconstruction informativeness and confidence with an annealed trade-off ($\beta$ shrinks as training proceeds). The current Gaussian field is rendered at the selected poses, refined by SDXL under four constraints---scene-specific LoRA fine-tuning, IP-Adapter appearance coherence, ControlNet depth structure (from Depth-Anything~V2), and img2img anchoring on the GS render---and then gated by a coverage and uncertainty check. Renderings with $>2\%$ uncovered area enter a SD2 inpainting branch before admission. The resulting pseudo-views feed the \emph{alternating GS optimization} (bottom): after a $1{,}000$-iteration warmup, training mixes pseudo-views (probability $0.25$, uncertainty-weighted loss) with ground-truth views (probability $0.75$, standard $L_1\!+\!\mathrm{SSIM}$ loss), under density-adaptive DropGaussian regularization.}
  \label{fig:pipeline}
\end{figure*}

Figure~\ref{fig:pipeline} illustrates the complete alternating optimization pipeline used by the proposed method.
\subsection{Preliminary}
\label{sec:preliminary}

\subsubsection{3D Gaussian Splatting}
\label{sec:prelim_3dgs}

3D Gaussian Splatting (3DGS)~\cite{kerbl20233dgs} represents a scene as a set of anisotropic 3D Gaussian primitives, typically initialized from a sparse Structure-from-Motion (SfM) point cloud. Each Gaussian primitive $i$ is parameterized by a center position $\boldsymbol{\mu}_i \in \mathbb{R}^3$, a full 3D covariance matrix $\boldsymbol{\Sigma}_i \in \mathbb{R}^{3 \times 3}$, an opacity value $\alpha_i \in [0,1]$, and a set of spherical harmonic (SH) coefficients for view-dependent color. The spatial extent of each primitive is defined by
\begin{equation}
  \mathcal{G}_i(\mathbf{x}) = \exp\!\left(-\frac{1}{2}(\mathbf{x} - \boldsymbol{\mu}_i)^\top \boldsymbol{\Sigma}_i^{-1} (\mathbf{x} - \boldsymbol{\mu}_i)\right).
  \label{eq:gaussian}
\end{equation}
To ensure positive semi-definiteness and enable unconstrained optimization, the covariance matrix is decomposed as $\boldsymbol{\Sigma}_i = \mathbf{R}_i \mathbf{S}_i \mathbf{S}_i^\top \mathbf{R}_i^\top$, where $\mathbf{R}_i$ is a rotation matrix derived from a unit quaternion $\mathbf{q}_i \in \mathbb{R}^4$ and $\mathbf{S}_i = \mathrm{diag}(\mathbf{s}_i)$ is a diagonal scaling matrix with $\mathbf{s}_i \in \mathbb{R}^3$.

Novel views are rendered by projecting the 3D Gaussians onto the image plane via a differentiable tile-based rasterizer. The Gaussians overlapping each pixel $\mathbf{p}$ are sorted by depth and composited front-to-back:
\begin{equation}
  C(\mathbf{p}) = \sum_{i \in \mathcal{N}} c_i \, \tilde{\alpha}_i \prod_{j=1}^{i-1}(1 - \tilde{\alpha}_j),
  \label{eq:alpha_blend}
\end{equation}
where $\mathcal{N}$ denotes the ordered set of Gaussians overlapping pixel $\mathbf{p}$, $c_i$ is the view-dependent color decoded from SH coefficients, and $\tilde{\alpha}_i = \alpha_i \, \mathcal{G}_i^{\mathrm{2D}}(\mathbf{p})$ is the effective opacity obtained by evaluating the projected 2D Gaussian at $\mathbf{p}$. Similarly, a depth map can be rendered by replacing the color $c_i$ with the distance from each Gaussian center to the camera:
\begin{equation}
  D(\mathbf{p}) = \sum_{i \in \mathcal{N}} \|\boldsymbol{\mu}_i - \mathbf{o}\|_2 \; \tilde{\alpha}_i \prod_{j=1}^{i-1}(1 - \tilde{\alpha}_j),
  \label{eq:depth_render}
\end{equation}
where $\mathbf{o}$ is the camera center. This rendered depth map plays a central role in our framework as the geometric conditioning signal for the diffusion model.

The Gaussian parameters are optimized via gradient descent with a photometric loss combining $\ell_1$ and structural dissimilarity:
\begin{equation}
  \mathcal{L}_{\mathrm{3DGS}} = (1 - \lambda)\,\mathcal{L}_1 + \lambda\,\mathcal{L}_{\mathrm{D\text{-}SSIM}}, \quad \lambda = 0.2.
  \label{eq:3dgs_loss}
\end{equation}
During training, an adaptive density control mechanism clones under-reconstructed Gaussians with large view-space positional gradients but small spatial extent, splits over-reconstructed ones with large extent, and periodically prunes near-transparent primitives. Under sparse input views, however, this densification receives insufficient gradient signal in unobserved regions, leaving large portions of the scene unrecovered.

\subsubsection{Latent Diffusion Models}
\label{sec:prelim_ldm}

Latent Diffusion Models (LDMs)~\cite{rombach2022ldm} perform the diffusion process in a compressed latent space to reduce computational cost. A pre-trained Variational Autoencoder (VAE) with encoder $\mathcal{E}$ and decoder $\mathcal{D}$ maps an image $\mathbf{x} \in \mathbb{R}^{H \times W \times 3}$ to a latent representation $\mathbf{z} = \mathcal{E}(\mathbf{x}) \in \mathbb{R}^{h \times w \times 4}$, where $h = H/8$ and $w = W/8$, and reconstructs it as $\hat{\mathbf{x}} = \mathcal{D}(\mathbf{z})$.

The forward diffusion process progressively adds Gaussian noise to the latent:
\begin{equation}
  q(\mathbf{z}_t \mid \mathbf{z}_0) = \mathcal{N}\!\left(\mathbf{z}_t;\, \sqrt{\bar{\alpha}_t}\,\mathbf{z}_0,\, (1 - \bar{\alpha}_t)\,\mathbf{I}\right),
  \label{eq:forward_diffusion}
\end{equation}
where $\bar{\alpha}_t = \prod_{s=1}^{t}(1 - \beta_s)$ is the cumulative product of the noise schedule. A UNet denoiser $\boldsymbol{\epsilon}_\theta$ is trained to predict the added noise $\boldsymbol{\epsilon}$ given the noisy latent $\mathbf{z}_t$, timestep $t$, and conditioning signal $\mathbf{c}$ (e.g., text prompt), by minimizing:
\begin{equation}
  \mathcal{L}_{\mathrm{LDM}} = \mathbb{E}_{\mathbf{z}_0,\,\boldsymbol{\epsilon} \sim \mathcal{N}(\mathbf{0},\mathbf{I}),\,t}\left[\left\|\boldsymbol{\epsilon} - \boldsymbol{\epsilon}_\theta(\mathbf{z}_t, t, \mathbf{c})\right\|_2^2\right].
  \label{eq:ldm_loss}
\end{equation}

Stable Diffusion XL (SDXL)~\cite{podell2024sdxl} extends this framework with a 2.6B-parameter UNet backbone and dual text encoders (CLIP ViT-L and OpenCLIP ViT-bigG), enabling richer text conditioning through cross-attention. SDXL further introduces micro-conditioning on the original image resolution and cropping parameters, improving high-resolution generation quality. In the \emph{img2img} setting, the reverse process starts from a partially noised version of a real image $\mathbf{z}_{t_0}$ with $t_0 < T$, rather than pure Gaussian noise, thereby preserving the coarse structure of the input while allowing the model to refine local details---a property that we exploit for pseudo-view generation.

\subsubsection{ControlNet}
\label{sec:prelim_controlnet}

ControlNet~\cite{zhang2023controlnet} augments a pre-trained diffusion model with spatially localized conditioning by creating a trainable copy of the UNet encoder while keeping the original weights frozen. Given a pre-trained network block $\mathbf{y} = \mathcal{F}(\mathbf{x};\,\Theta)$, ControlNet locks $\Theta$ and introduces a trainable copy with parameters $\Theta_c$, connected through \emph{zero convolution} layers---$1\!\times\!1$ convolutions initialized with zero weights and biases. The output becomes:
\begin{equation}
  \mathbf{y}_c = \mathcal{F}(\mathbf{x};\,\Theta) + \mathcal{Z}\!\left(\mathcal{F}\!\left(\mathbf{x} + \mathcal{Z}(\mathbf{c}_f;\,\Theta_{z1});\,\Theta_c\right);\,\Theta_{z2}\right),
  \label{eq:controlnet}
\end{equation}
where $\mathbf{c}_f$ is the external conditioning signal (e.g., a depth map), and $\Theta_{z1}$, $\Theta_{z2}$ are the parameters of the two zero convolution layers. Because zero convolutions produce zero outputs at initialization, the ControlNet branch contributes nothing before training begins, ensuring stable integration with the pre-trained model. In practice, a trainable copy of the UNet encoder blocks is created, and the ControlNet residuals are added to the corresponding decoder skip connections. In our framework, the conditioning signal $\mathbf{c}_f$ is a depth map rendered from the current Gaussian field (Eq.~\ref{eq:depth_render}), which guides the diffusion model to generate images that respect the reconstructed scene geometry.

\subsubsection{DropGaussian}
\label{sec:prelim_dropgaussian}

Dropout~\cite{srivastava2014dropout} is a widely used regularization technique in neural networks that randomly deactivates a fraction of neurons during training, preventing co-adaptation and improving generalization. We extend this idea to the 3DGS representation by applying stochastic opacity dropout to the Gaussian primitives. At each training iteration, a subset of Gaussians are randomly selected and have their opacity set to zero, effectively removing them from the rendered image. The remaining primitives must independently produce a consistent rendering without relying on specific neighbors.

Concretely, given a drop ratio $r_{\mathrm{drop}} \in [0, 1)$, a Bernoulli mask $\mathbf{m} \in \{0, 1\}^{N_{\mathrm{pts}}}$ is sampled with $P(m_i = 0) = r_{\mathrm{drop}}$. The effective opacity for rendering becomes:
\begin{equation}
  \alpha_i' = \alpha_i \cdot \frac{m_i}{1 - r_{\mathrm{drop}}},
  \label{eq:dropgaussian}
\end{equation}
where the $(1 - r_{\mathrm{drop}})^{-1}$ compensation ensures that the expected rendered pixel value remains unbiased: $\mathbb{E}[\alpha_i'] = \alpha_i$. Only the opacity channel is masked; the gradients for position, covariance, and color flow through all primitives regardless of the dropout state, so that geometric parameters continue to receive optimization signal even when a primitive is dropped.

\subsection{Framework Overview}
\label{sec:overview}

\paragraph{Problem definition}
Given $N$ sparse input images $\mathcal{I} = \{I_1, \ldots, I_N\}$ of a scene, we run COLMAP~\cite{schoenberger2016sfm} Structure-from-Motion and Multi-View Stereo (MVS) on these $N$ images alone to obtain camera poses $\mathcal{P} = \{P_1, \ldots, P_N\}$ and a dense initial point cloud. The Gaussian field $\boldsymbol{\theta}$ is initialized from this MVS point cloud and optimized to render high-quality novel views. Because the sparse observations are severely under-constrained, we supplement the training set with $\tilde{N}$ pseudo-views $\tilde{\mathcal{I}} = \{\tilde{I}_1, \ldots, \tilde{I}_{\tilde{N}}\}$ generated by a scene-adapted diffusion model at novel poses $\tilde{\mathcal{P}} = \{\tilde{P}_1, \ldots, \tilde{P}_{\tilde{N}}\}$, which are interpolated or extrapolated from the SfM poses.

\paragraph{Alternating optimization}
The reconstruction is cast as a bi-level optimization problem. At the outer level, a diffusion model parameterized by $\Theta$ is adapted to the scene and used to hallucinate pseudo-views; at the inner level, the Gaussian field $\boldsymbol{\theta}$ is optimized under supervision from both ground-truth and pseudo-views:
\begin{equation}
  \begin{split}
  \boldsymbol{\theta}^* &= \arg\min_{\boldsymbol{\theta}}\; \sum_{i=1}^{N}\mathcal{L}_{\mathrm{3DGS}}\!\left(R(P_i;\boldsymbol{\theta}),\, I_i\right) \\
  &\quad+ \sum_{j=1}^{\tilde{N}} \mathcal{L}_{\mathrm{pseudo}}\!\left(R(\tilde{P}_j;\boldsymbol{\theta}),\, \tilde{I}_j\right),
  \end{split}
  \label{eq:bilevel_outer}
\end{equation}
\begin{equation}
  \mathrm{s.t.}\quad \Theta^* = \arg\min_{\Theta}\; \sum_{i=1}^{N}\mathcal{L}_{\mathrm{LDM}}(I_i,\, D_i;\, \Theta),
  \label{eq:bilevel_inner}
\end{equation}
where $R(\cdot;\boldsymbol{\theta})$ denotes the differentiable Gaussian rasterizer, $D_i$ is the depth map for view $i$, and $\tilde{I}_j = \mathrm{sample}[\boldsymbol{\epsilon}_{\Theta^*}(\tilde{P}_j)]$ is a pseudo-view sampled from the adapted diffusion model. In practice, the two levels are solved by alternation: the Gaussian field is optimized for a fixed number of iterations, after which the diffusion model generates a new batch of pseudo-views that are fed back as additional supervision.

\paragraph{Training pipeline}
The complete training process consists of three stages:

\emph{Stage~I: Warmup} (iterations $1$ to $T_w$, default $T_w\!=\!1000$). The Gaussian field is trained exclusively on the $N$ ground-truth views using $\mathcal{L}_{\mathrm{3DGS}}$ (Eq.~\ref{eq:3dgs_loss}), with standard adaptive densification. This establishes a coarse scene geometry before introducing diffusion-generated supervision.

\emph{Stage~II: Diffusion phase} (triggered every $\Delta T\!=\!500$ GS iterations). Each diffusion phase performs the following sequence of operations:
\begin{enumerate}
  \item \textbf{View selection:} GAPS (Section~\ref{sec:gaps}) selects $n_{\mathrm{active}}$ candidate poses that jointly maximize reconstruction informativeness and generative reliability.
  \item \textbf{Depth estimation:} For each ground-truth view, a monocular depth estimator (Depth-Anything V2~\cite{yang2024depthanythingv2}) produces depth maps as ControlNet conditioning. These are cached and reused across phases.
  \item \textbf{Scene adaptation:} In the first diffusion phase, the SDXL UNet is fine-tuned via LoRA (Section~\ref{sec:lora}) on the ground-truth images for 50 gradient steps. The adapted weights are frozen and reused in all subsequent phases.
  \item \textbf{Pseudo-view rendering:} The current Gaussian field is rendered at each selected pose to obtain initial images, depth maps, and alpha coverage maps.
  \item \textbf{Constrained generation:} The four-fold constrained SDXL pipeline (Section~\ref{sec:generation}) generates hallucinated images from the Gaussian renderings. Views with substantial uncovered regions undergo a second inpainting stage (Section~\ref{sec:two_stage}).
  \item \textbf{Admission and uncertainty:} Each generated pseudo-view is evaluated against dual admission criteria (Section~\ref{sec:dual_admission}). Admitted views are paired with per-pixel uncertainty maps for subsequent confidence-weighted training.
\end{enumerate}

\emph{Stage~III: GS optimization with pseudo-views} (continuous between diffusion phases). At each iteration, a training view is sampled: with probability $p_{\mathrm{pseudo}}\!=\!0.25$ an admitted pseudo-view is chosen and trained with the uncertainty-weighted loss $\mathcal{L}_{\mathrm{pseudo}}$ (Eq.~\ref{eq:pseudo_loss}); otherwise a ground-truth view is used with $\mathcal{L}_{\mathrm{3DGS}}$. Adaptive densification operates only on ground-truth iterations to prevent pseudo-view noise from corrupting the point cloud topology. Density-adaptive DropGaussian regularization (Section~\ref{sec:dropgaussian}) is applied throughout to mitigate overfitting.

Stages~II and~III alternate until the total iteration budget is exhausted ($T\!=\!10{,}000$ by default), yielding approximately 18 diffusion phases. As training progresses, the improving Gaussian field produces better depth maps and renderings, which in turn improve pseudo-view quality---creating a virtuous cycle of mutual refinement between the 3D representation and the generative model.

\subsection{Four-fold Constrained Pseudo-view Generation}
\label{sec:generation}

A central challenge in diffusion-assisted sparse-view reconstruction is constraining the generative model so that its outputs are geometrically and stylistically consistent with the target scene, rather than plausible but arbitrary images. We address this by channeling the generation through four complementary conditioning mechanisms---ControlNet for geometric structure, IP-Adapter for appearance coherence, LoRA for scene-specific memory, and img2img denoising for structural anchoring---each restricting a different axis of the diffusion model's output space. For regions where the Gaussian field provides no coverage, we further employ a dedicated inpainting stage.

\paragraph{Depth-conditioned generation via ControlNet}
\label{sec:controlnet_depth}

As described in Section~\ref{sec:prelim_controlnet}, ControlNet injects spatial conditioning into the diffusion process. In our pipeline, the conditioning signal $\mathbf{c}_f$ is a depth map rendered from the current Gaussian field at the target pseudo-view pose via Eq.~\eqref{eq:depth_render}. To prevent unreliable depth from misleading the generation, we apply a reliability mask: pixels with alpha coverage below a threshold $\tau_\alpha$ or with high local depth variance are zeroed out before being passed to ControlNet. Additionally, the conditioning scale is adapted to the view's coverage:
\begin{equation}
  s_{\mathrm{cn}}(v) = \bar{s}_{\mathrm{cn}} \cdot \min\!\left(1,\; \frac{\mathrm{coverage}(v)}{\tau_{\mathrm{clip}}}\right),
  \label{eq:adaptive_cn}
\end{equation}
where $\bar{s}_{\mathrm{cn}} = 1.0$ is the base conditioning scale and $\tau_{\mathrm{clip}} = 0.3$ is the coverage threshold for full-strength conditioning. This allows the model to hallucinate more freely in low-coverage regions where depth estimates are unreliable.

\paragraph{Appearance guidance via IP-Adapter}
\label{sec:ipadapter}

While ControlNet constrains geometry, it provides no guidance on appearance. IP-Adapter~\cite{ye2023ipadapter} addresses this by injecting a visual style reference into the UNet's cross-attention layers. Given a reference image $I_{\mathrm{ref}}$ (the nearest ground-truth view by pose distance), a pre-trained CLIP ViT-H/14 image encoder extracts a style embedding $\mathbf{e}_{\mathrm{img}} \in \mathbb{R}^{1024}$. This embedding is projected into key--value pairs and attended to alongside the text conditioning in each cross-attention layer:
\begin{equation}
  \mathbf{Z} = \mathrm{Attn}(\mathbf{Q},\, \mathbf{K}_t,\, \mathbf{V}_t) + \lambda_{\mathrm{ip}} \cdot \mathrm{Attn}(\mathbf{Q},\, \mathbf{K}_{\mathrm{img}},\, \mathbf{V}_{\mathrm{img}}),
  \label{eq:ipadapter}
\end{equation}
where $\mathbf{K}_t, \mathbf{V}_t$ are the text-conditioned keys and values, $\mathbf{K}_{\mathrm{img}}, \mathbf{V}_{\mathrm{img}}$ are projected from $\mathbf{e}_{\mathrm{img}}$, and $\lambda_{\mathrm{ip}} = 0.5$ controls the strength of the appearance guidance. This mechanism transfers color palette, lighting, and material characteristics from the observed views without imposing pixel-level constraints that would conflict with the novel viewpoint.

\paragraph{Scene-specific adaptation via LoRA}
\label{sec:lora}

Pre-trained diffusion models encode generic image priors but lack knowledge of the specific scene being reconstructed. We bridge this gap through Low-Rank Adaptation (LoRA)~\cite{hu2022lora}, which fine-tunes the UNet on the available ground-truth images. LoRA injects trainable low-rank matrices into the cross-attention layers:
\begin{equation}
  \mathbf{W}' = \mathbf{W}_0 + \mathbf{B}\mathbf{A}, \quad \mathbf{B} \in \mathbb{R}^{d \times r},\; \mathbf{A} \in \mathbb{R}^{r \times k},
  \label{eq:lora}
\end{equation}
where $\mathbf{W}_0$ is the frozen pre-trained weight, $r = 4 \ll \min(d, k)$ is the rank, and $\mathbf{B}, \mathbf{A}$ are trained on the ground-truth views by minimizing the denoising objective (Eq.~\ref{eq:ldm_loss}) with ControlNet depth conditioning active. The adaptation targets the query, key, value, and output projection matrices (\texttt{to\_q}, \texttt{to\_k}, \texttt{to\_v}, \texttt{to\_out}) in all cross-attention blocks, adding approximately 0.06\% trainable parameters relative to the full UNet. LoRA is trained once during the first diffusion phase for 50 gradient steps and reused in all subsequent phases, encoding a compact scene-specific memory that steers the generation toward the target scene's textures and materials.

\paragraph{Structural anchoring via img2img}
\label{sec:img2img}

The three mechanisms above operate on the conditioning side of the diffusion process. We further constrain the \emph{initialization} by using the img2img setting: instead of starting from pure noise, the reverse process begins from a partially noised version of the current Gaussian rendering at the target pose. The denoising strength---which controls the fraction of diffusion steps actually executed---is adapted to the view's Gaussian coverage:
\begin{equation}
  \sigma(v) = \sigma_{\max} - \mathrm{coverage}(v) \cdot (\sigma_{\max} - \sigma_{\min}),
  \label{eq:adaptive_strength}
\end{equation}
where $\sigma_{\min} = 0.3$ and $\sigma_{\max} = 0.85$. For well-covered views, low denoising strength preserves the Gaussian rendering's structure while refining details; for poorly covered views, higher strength allows the model to hallucinate content more freely. This adaptive scheme prevents the diffusion model from either over-modifying reliable regions or under-correcting deficient ones.

\paragraph{Two-stage generation for incomplete views}
\label{sec:two_stage}

When a pseudo-view contains substantial regions with no Gaussian coverage (detected as pixels where the rendered depth $D(\mathbf{p}) < \tau_d$), a two-stage pipeline is employed:
\begin{enumerate}
  \item \textbf{Stage~1 (SDXL img2img):} The full four-constraint SDXL pipeline generates an initial image $I_{\mathrm{s1}}$ from the Gaussian rendering, as described above. This stage produces high-quality content in well-covered regions but may leave artifacts in areas with no Gaussian support.
  \item \textbf{Stage~2 (SD2 inpainting):} A pre-trained Stable Diffusion 2 inpainting model receives $I_{\mathrm{s1}}$ together with a binary mask $\mathbf{M}$ identifying uncovered regions (pixels where $D(\mathbf{p}) < \tau_d$, dilated by a $15\!\times\!15$ morphological kernel to ensure seamless blending). The inpainting model regenerates only the masked pixels while preserving the Stage~1 content elsewhere.
\end{enumerate}
The final pseudo-view is composited as:
\begin{equation}
  I_{\mathrm{pseudo}}(\mathbf{p}) = \mathbf{M}(\mathbf{p})\, I_{\mathrm{s2}}(\mathbf{p}) + (1 - \mathbf{M}(\mathbf{p}))\, I_{\mathrm{s1}}(\mathbf{p}),
  \label{eq:two_stage}
\end{equation}
where $I_{\mathrm{s2}}$ is the inpainting output. This design delegates geometrically grounded regions to the fully constrained SDXL pipeline and uncovered regions to a dedicated inpainting model, avoiding the need for a single model to handle both tasks simultaneously.

\subsection{Generative Active Pseudo-view Selection}
\label{sec:gaps}

Existing diffusion-assisted methods place pseudo-view cameras at fixed interpolations between training pairs~\cite{kong2025gsgs}, without considering which viewpoints would most benefit the reconstruction or where the diffusion model can generate reliably. We propose \emph{Generative Active Pseudo-view Selection} (GAPS), a scoring framework that jointly balances reconstruction informativeness and generative reliability.

\paragraph{Candidate pool construction}
\label{sec:gaps_candidates}

For each pair of training cameras $(P_i, P_j)$, we generate candidate poses by interpolating and extrapolating along the geodesic connecting them. Rotations are interpolated via SLERP on the rotation manifold:
\begin{equation}
  \mathbf{R}(\alpha) = \mathbf{R}_i \cdot \left(\mathbf{R}_i^{-1}\mathbf{R}_j\right)^\alpha, \quad
  \mathbf{t}(\alpha) = (1 - \alpha)\,\mathbf{t}_i + \alpha\,\mathbf{t}_j,
  \label{eq:slerp}
\end{equation}
where $\alpha \in [a_{\min}, a_{\max}]$ controls the position along (and beyond) the baseline. We sample candidates at uniform $\alpha$ intervals, and define the extrapolation distance as $d_{\mathrm{extrap}} = \max(0, -\alpha) + \max(0, \alpha - 1)$, which is zero for interpolated views and increases for extrapolated ones.

To avoid overly aggressive extrapolation early in training when the Gaussian field is still coarse, we employ a \emph{progressive tier} strategy that gradually widens the $\alpha$ range over the course of training. The range begins conservatively near the interpolation interval and expands outward in discrete stages as the reconstruction matures. Tier upgrades are triggered when the mean uncertainty variance of currently admitted pseudo-views falls below a threshold, providing a data-driven signal that the model can reliably handle more distant viewpoints.

\paragraph{Scoring function}
\label{sec:gaps_score}

Each candidate view $v$ receives a composite score that trades off reconstruction gain against generative reliability:
\begin{equation}
  \mathrm{score}(v) = I_{\mathrm{recon}}(v) \cdot C_{\mathrm{gen}}(v)^{\beta},
  \label{eq:gaps_score}
\end{equation}
where $I_{\mathrm{recon}}$ quantifies how much new information the view would contribute and $C_{\mathrm{gen}}$ estimates how reliably the diffusion model can generate it.

\paragraph{Reconstruction informativeness}
We combine three complementary signals:
\begin{equation}
  I_{\mathrm{recon}}(v) = w_s\,(1 - \bar{\alpha}_v) + w_e\,H_v + w_n\,\eta_v,
  \label{eq:i_recon}
\end{equation}
where $\bar{\alpha}_v$ is the mean alpha coverage rendered from the current Gaussian field (its complement measures sparsity), $H_v$ is the mean local entropy of the rendered image computed over a $3\!\times\!3$ sliding window (measuring rendering instability), and $\eta_v = \min\!\left(1,\, 4 d_{\min}/(1 + d_{\min})\right)$ is a novelty term based on the minimum pose distance $d_{\min}$ to all supervised views. The weights $w_s = w_e = 0.2$ and $w_n = 0.6$ emphasize viewpoint novelty as the primary driver of reconstruction gain.

\paragraph{Generative confidence}
The confidence score blends a model-based estimate with historical reliability:
\begin{equation}
  C_{\mathrm{gen}}(v) = (1 - \gamma)\,C_{\mathrm{model}}(v) + \gamma\,C_{\mathrm{hist}}(v),
  \label{eq:c_gen}
\end{equation}
where $\gamma = 0.7$ weights history heavily after the first few diffusion phases. The model-based component $C_{\mathrm{model}} = w_{\mathrm{init}}\,C_{\mathrm{init}} + w_{\mathrm{ref}}\,C_{\mathrm{ref}}$ combines an initial quality estimate (based on Gaussian coverage and render contrast) with a reference accessibility term $C_{\mathrm{ref}} = \exp(-d_{\min}/0.5)$ that decays with distance from the nearest ground-truth view. The historical component $C_{\mathrm{hist}} = 1 - \min(1,\, \bar{\sigma}_v^2 / 0.3)$ converts the mean uncertainty variance from prior phases into a confidence value.

\begin{figure}[t]
  \centering
  \includegraphics[width=\linewidth]{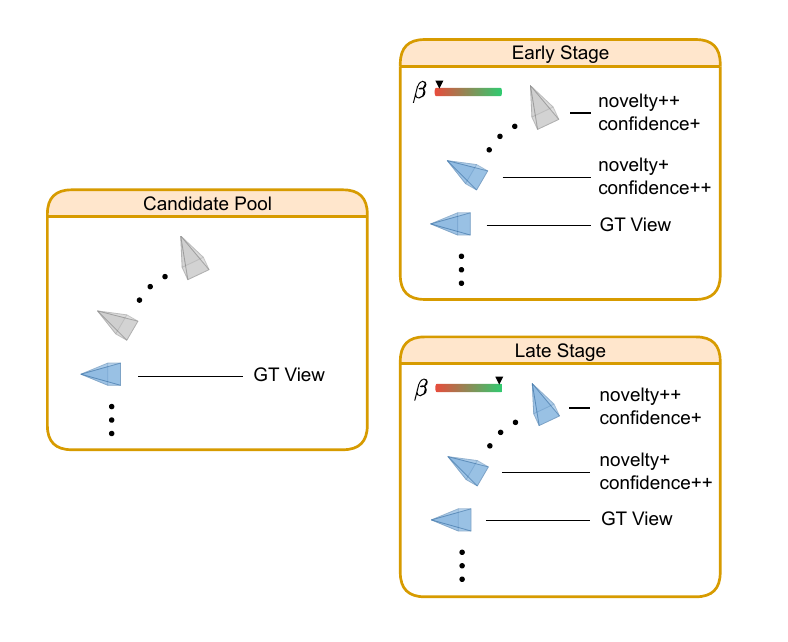}
  \caption{Illustration of GAPS $\beta$-annealing. \textbf{Left}: candidate pool generated by interpolation and extrapolation between GT views. \textbf{Middle}: early-stage selection favors high-confidence views near GT cameras ($\beta$ large). \textbf{Right}: late-stage selection shifts toward novel viewpoints with higher reconstruction gain ($\beta$ small).}
  \label{fig:gaps}
\end{figure}

\paragraph{Annealing schedule}
The exponent $\beta$ controls the influence of generative confidence on the final score. We anneal it linearly over training:
\begin{equation}
  \beta(t) = \beta_{\mathrm{start}} + (\beta_{\mathrm{end}} - \beta_{\mathrm{start}}) \cdot \frac{t}{T},
  \label{eq:beta_anneal}
\end{equation}
where $\beta_{\mathrm{start}} > \beta_{\mathrm{end}}$. In early stages, the high exponent strongly penalizes low-confidence views, favoring conservative near-GT viewpoints where diffusion is most reliable. As the Gaussian field improves and historical confidence accumulates, the reduced exponent shifts priority toward novel, informative viewpoints even if their generative confidence is moderate (see Fig.~\ref{fig:gaps}).

\paragraph{Greedy selection with diversity}
\label{sec:gaps_selection}

From the scored candidate pool, we greedily select $n_{\mathrm{active}}$ views (default 16) by iteratively choosing the highest-scoring candidate and removing all candidates within a minimum pose diversity distance $d_{\mathrm{div}} = 0.15$. We further enforce a per-pair cap to prevent clustering: no camera pair may contribute more than $\lceil n_{\mathrm{active}} / n_{\mathrm{pairs}} \rceil$ selected views. The pose distance between two views is defined as
\begin{equation}
  d(v_1, v_2) = \|\mathbf{t}_1 - \mathbf{t}_2\| + 0.5 \cdot \arccos\frac{\mathrm{tr}(\mathbf{R}_1^\top \mathbf{R}_2) - 1}{2},
  \label{eq:pose_dist}
\end{equation}
combining translational and rotational components.

\subsection{Density-adaptive DropGaussian for Pseudo-view Training}
\label{sec:dropgaussian}

With only three ground-truth views, clusters of overlapping Gaussians tend to co-adapt: each primitive memorizes a narrow angular slice that, composited with its neighbors, reproduces the training views but fails to generalize. This directly undermines pseudo-view supervision, as gradient signal from novel viewpoints is absorbed by fragile clusters rather than individually robust primitives. We integrate DropGaussian (Section~\ref{sec:prelim_dropgaussian}) into the alternating framework with two key adaptations.

First, dropout is applied during GS optimization (Stage~III) on both ground-truth and pseudo-view iterations, forcing each surviving Gaussian to independently account for novel viewpoints. It is deliberately \emph{disabled} during the diffusion phase rendering (Stage~II), where full-fidelity renders are needed for ControlNet depth conditioning, img2img anchoring, and the alpha coverage maps that drive pseudo-view admission (Section~\ref{sec:dual_admission}) and GAPS scoring (Section~\ref{sec:gaps}).

Second, we make the drop ratio density-adaptive:
\begin{equation}
  r_{\mathrm{drop}}(t) = r_{\max} \cdot \frac{t}{T} \cdot \min\!\left(\frac{N_{\mathrm{pts}}}{N_{\mathrm{thr}}},\; 1\right),
  \label{eq:drop_ratio}
\end{equation}
where $r_{\max} = 0.2$, $N_{\mathrm{pts}}$ is the current Gaussian count, and $N_{\mathrm{thr}} = 200{,}000$. This creates a self-regulating feedback loop: as pseudo-view supervision expands the reconstructed volume and densification grows the point cloud, the increasing $N_{\mathrm{pts}}$ automatically strengthens regularization, counteracting the elevated overfitting risk that accompanies field expansion. The temporal ramp further coordinates with the alternating schedule---weak dropout during early phases where pseudo-views provide the largest information gain, stronger in later phases where memorizing redundant content becomes the dominant risk. Together with the pixel-level uncertainty weighting (Section~\ref{sec:unc_loss}), this forms a two-scale quality control: uncertainty weighting modulates \emph{where} pseudo-view gradients are trusted on the image plane, while DropGaussian modulates \emph{how} they are distributed across 3D primitives.

\subsection{Pseudo-view Admission and Uncertainty-weighted Loss}
\label{sec:admission}

Not all generated pseudo-views contribute positively to training; low-quality generations can introduce artifacts and destabilize optimization. We address this through a dual-criterion admission gate followed by an uncertainty-weighted loss that modulates the contribution of each admitted pseudo-view at the pixel level.

\paragraph{Dual-criterion admission}
\label{sec:dual_admission}

A generated pseudo-view is admitted to the training set only if it satisfies both of the following conditions:
\begin{enumerate}
  \item \textbf{Geometric coverage}: the mean alpha map $\bar{\alpha}_v$, rendered from the Gaussian field at the pseudo-view pose, must exceed a minimum threshold: $\bar{\alpha}_v \geq \tau_{\mathrm{cov}}$ (default $0.25$). This rejects views where the Gaussian field has insufficient content to anchor the generation.
  \item \textbf{Generation consistency}: the mean per-pixel residual between the hallucinated image $I_{\mathrm{hall}}$ and the Gaussian rendering $I_{\mathrm{GS}}$ must be below a threshold: $\frac{1}{|\mathcal{P}|}\sum_{\mathbf{p}} |I_{\mathrm{hall}}(\mathbf{p}) - I_{\mathrm{GS}}(\mathbf{p})| < \tau_{\mathrm{res}}$ (default $0.7$). This filters out generations that deviate excessively from the current reconstruction.
\end{enumerate}
Views that fail either criterion are excluded from the training set but remain in the candidate pool for re-evaluation in future diffusion phases.

\paragraph{Uncertainty-weighted loss}
\label{sec:unc_loss}

For each admitted pseudo-view, we compute a per-pixel uncertainty map as the absolute residual between the hallucinated and rendered images:
\begin{equation}
  u(\mathbf{p}) = \left|I_{\mathrm{hall}}(\mathbf{p}) - I_{\mathrm{GS}}(\mathbf{p})\right|.
  \label{eq:uncertainty}
\end{equation}
This residual is high in regions where the diffusion model deviates from the Gaussian rendering---either because the generation is unreliable or because the rendering is incomplete. We convert this into a per-pixel confidence weight:
\begin{equation}
  w(\mathbf{p}) = \alpha(\mathbf{p}) \cdot \left(1 - \lambda_u \cdot \frac{u(\mathbf{p})}{\max_{\mathbf{p}'} u(\mathbf{p}') + \epsilon}\right),
  \label{eq:confidence_weight}
\end{equation}
where $\alpha(\mathbf{p})$ is the rendered alpha coverage at pixel $\mathbf{p}$, $\lambda_u = 0.7$ controls the uncertainty downweighting strength, and $\epsilon$ is a small constant for numerical stability. This weight is low for pixels with either low Gaussian coverage (unreliable rendering) or high residual (unreliable generation), concentrating the learning signal on regions where both the rendering and the generation agree.

For views obtained by extrapolation beyond the training cameras, we apply an additional distance-based decay:
\begin{equation}
  w_{\mathrm{extrap}}(v) = \exp\!\left(-\kappa \cdot d_{\mathrm{extrap}}(v)\right),
  \label{eq:extrap_decay}
\end{equation}
with $\kappa = 2.0$, which attenuates the contribution of distant extrapolated views where generative accuracy naturally degrades.

The full pseudo-view training loss for a single view $v$ is:
\begin{equation}
  \begin{split}
  \mathcal{L}_{\mathrm{pseudo}}(v) &= w_{\mathrm{PLW}} \cdot w_{\mathrm{extrap}}(v) \\
  &\quad\cdot \sum_{\mathbf{p}} w(\mathbf{p})\Big[(1\!-\!\lambda)\,\lvert I_v(\mathbf{p}) - \hat{I}_v(\mathbf{p})\rvert \\
  &\qquad\qquad\quad + \lambda\,(1 - \mathrm{SSIM}_{\mathbf{p}})\Big],
  \end{split}
  \label{eq:pseudo_loss}
\end{equation}
where $w_{\mathrm{PLW}} = 0.35$ is the global pseudo-loss weight that balances pseudo-view supervision against the ground-truth loss (Eq.~\ref{eq:3dgs_loss}), $I_v$ is the Gaussian rendering, and $\hat{I}_v$ is the hallucinated pseudo-view. During training, each iteration samples either a ground-truth view (with probability $1 - p_{\mathrm{pseudo}}$) or an admitted pseudo-view (with probability $p_{\mathrm{pseudo}} = 0.25$), applying the corresponding loss function.

\paragraph{Mitigating admission saturation}
\label{sec:force_new}

As training progresses and the Gaussian field converges, the admission criterion may become saturated: newly generated pseudo-views closely match the existing rendering, yielding near-zero residuals but contributing little new information. To counteract this, every $k_{\mathrm{new}}$ diffusion phases (default 3), we replace a fraction $\rho_{\mathrm{new}} = 0.25$ of the currently selected views with previously unadmitted candidates. This forced exploration injects fresh viewpoints into the training set, breaking potential echo-chamber effects where the diffusion model merely reproduces what the Gaussian field already knows.

\section{Experiments}
\label{sec:experiments}
\subsection{Datasets and Metrics}
\label{sec:datasets}

We evaluate on two widely used benchmarks for novel view synthesis. \textbf{LLFF}~\cite{mildenhall2019llff} contains eight forward-facing real-world scenes (\textit{fern}, \textit{flower}, \textit{fortress}, \textit{horns}, \textit{leaves}, \textit{orchids}, \textit{room}, \textit{trex}). \textbf{Mip-NeRF 360}~\cite{barron2022mipnerf360} comprises seven unbounded scenes with complex geometry, including outdoor (\textit{bicycle}, \textit{garden}, \textit{stump}) and indoor (\textit{bonsai}, \textit{counter}, \textit{kitchen}, \textit{room}) cases. Images on both datasets are downsampled to eighth resolution (\texttt{-r 8}) and we hold out every 8th image for testing (\texttt{llffhold=8}), uniformly sampling $N$ training views from the remainder with $N\in\{3,6,9\}$ on LLFF and $N\in\{12,24\}$ on Mip-NeRF 360. Reconstruction quality is reported in PSNR, SSIM~\cite{wang2004ssim} and LPIPS~\cite{zhang2018lpips}. We compute LPIPS with the VGG-backbone \texttt{lpipsPyTorch} implementation shared by recent sparse-view 3DGS work~\cite{zhu2024fsgs, zhang2024cor_gs, park2025dropgaussian} so that our LPIPS values are directly comparable to theirs on the same resolution protocol.

\subsection{Baseline}
\label{sec:baselines}

We compare against vanilla 3DGS~\cite{kerbl20233dgs} as the foundational baseline. The baseline and our method share the same sparse MVS initialization (Section~\ref{sec:exp_init}), so any performance difference reflects the reconstruction algorithm rather than the input point cloud.

\subsection{Implementation Details}
\label{sec:impl}

All experiments are conducted on a single NVIDIA RTX~4090D GPU (24~GB). Our implementation follows the alternating training schedule and four-fold constrained generation pipeline described in Section~\ref{sec:method}, and the 3DGS optimizer (learning rates, opacity reset interval, D-SSIM weight) follows the original 3DGS implementation~\cite{kerbl20233dgs}. Each scene is trained for $10{,}000$ iterations. The GAPS pool is sized to the dataset: $16$ active pseudo-views per diffusion phase on LLFF and $24$ on Mip-NeRF~360 to match the wider $360^\circ$ viewpoint range.

\paragraph{Point cloud initialization}
\label{sec:exp_init}
We adopt the sparse-view COLMAP+MVS initialization that is now the de facto standard in recent sparse-view 3DGS work~\cite{zhu2024fsgs, zhang2024cor_gs, park2025dropgaussian}: given the $N$ sampled training images, we run COLMAP~\cite{schoenberger2016sfm} \texttt{feature\_extractor} and exhaustive \texttt{matcher} on those $N$ images alone, then invoke \texttt{point\_triangulator} with the ground-truth poses (no pose re-estimation), followed by \texttt{image\_undistorter}, \texttt{patch\_match\_stereo}, and \texttt{stereo\_fusion} to obtain a dense fused point cloud. This differs from the original single-scene 3DGS~\cite{kerbl20233dgs} evaluation protocol, which ingests a COLMAP reconstruction built over all scene images: here the reconstruction only sees the $N$ training views, so no information from the held-out test split leaks into the point cloud. The same point cloud is supplied to vanilla 3DGS and our method via the \texttt{--init\_ply} interface.

\section{Results and Discussion}
\label{sec:results}
\subsection{Qualitative Comparison}
\label{sec:qualitative}

We first visualize held-out test renders on representative scenes from both datasets, comparing our reconstructions against vanilla 3DGS~\cite{kerbl20233dgs} under the same sparse MVS initialization (Section~\ref{sec:exp_init}) and the same training budget. Per-image PSNR and SSIM values are inset in each tile for reference; aggregate quantitative results are reported in Section~\ref{sec:quantitative}.

\begin{figure}[!t]
  \centering
  \begin{subfigure}{\linewidth}
    \centering
    \includegraphics[width=\linewidth]{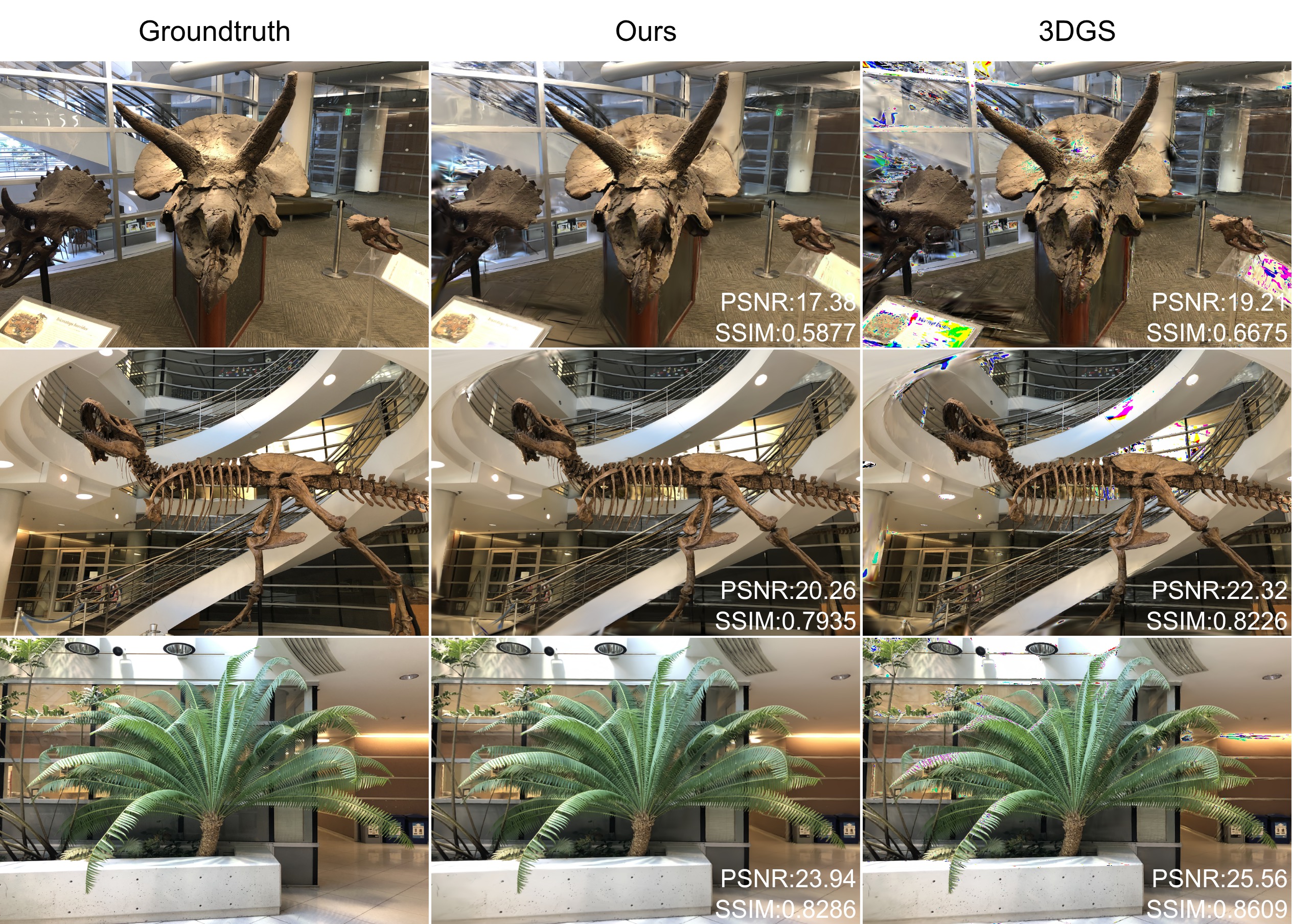}
    \caption{LLFF (rows: \emph{horns}, \emph{trex}, \emph{fern}).}
    \label{fig:qual_llff}
  \end{subfigure}

  \vspace{0.2em}

  \begin{subfigure}{\linewidth}
    \centering
    \includegraphics[width=\linewidth]{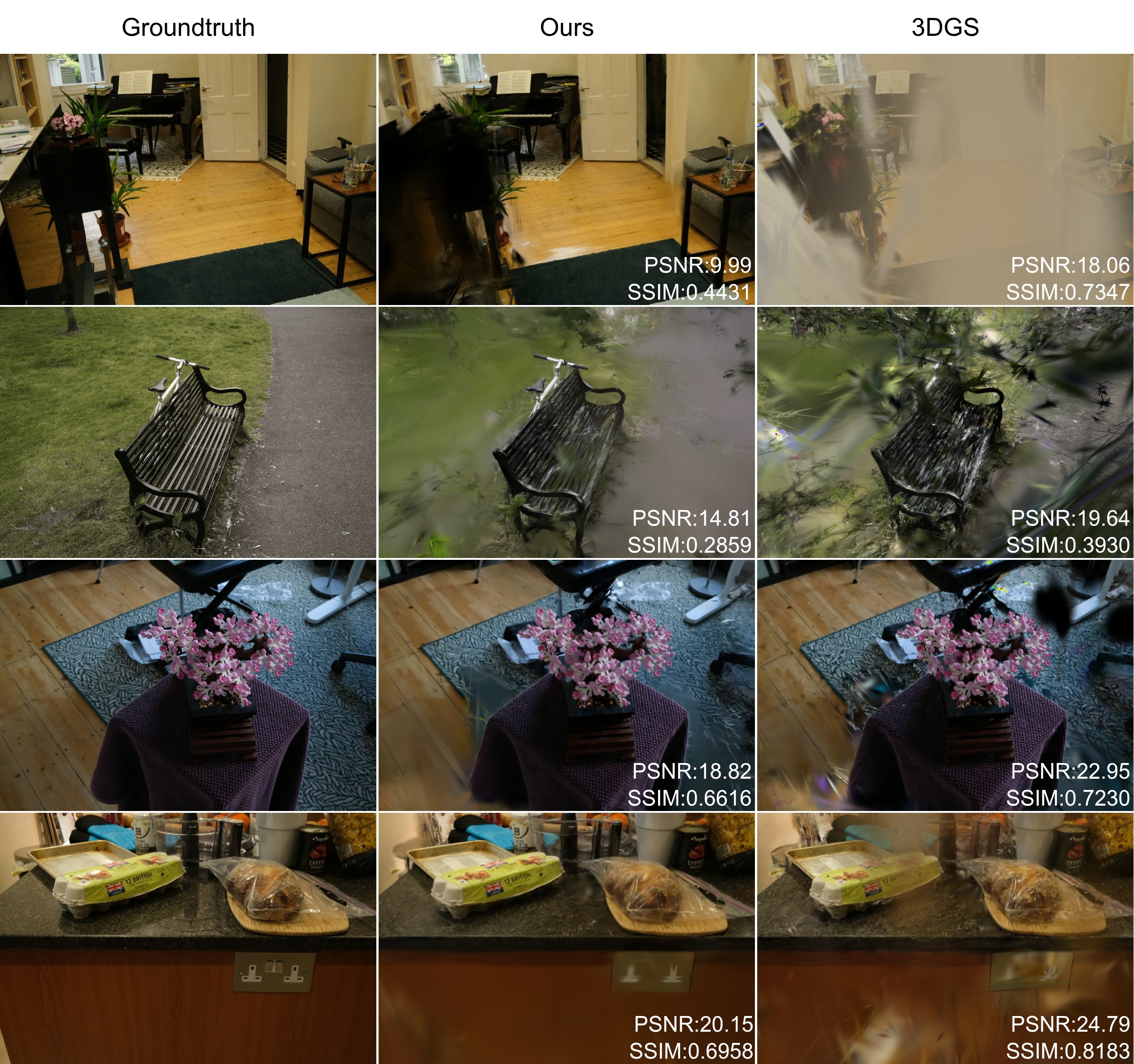}
    \caption{Mip-NeRF 360 (rows: \emph{room}, \emph{garden}, \emph{bonsai}, \emph{counter}).}
    \label{fig:qual_360}
  \end{subfigure}

  \caption{Qualitative comparison on held-out test views. Each panel shows the ground truth (left), our reconstruction (middle), and vanilla 3DGS~\cite{kerbl20233dgs} (right) trained from the same sparse input set and MVS initialization.}
  \label{fig:qualitative}
\end{figure}

\paragraph{LLFF}
Figure~\ref{fig:qual_llff} reports test renders on three forward-facing LLFF scenes that stress different failure modes of sparse-view reconstruction: thin rigid structures (\emph{horns}), repeated articulated geometry under strong occlusion (\emph{trex}), and dense high-frequency texture (\emph{fern}). With only a handful of input views, vanilla 3DGS lacks the gradient signal to populate the unobserved volume between training cameras, and the resulting renders show characteristic artifacts---ghosted edges around the triceratops skull, broken bone structure on the \emph{trex} ribcage, and grainy speckle across the \emph{fern} fronds. Our diffusion-supervised pseudo-views inject geometrically consistent content into exactly these gaps, and the corresponding renders preserve both the global silhouette of the foreground objects and the fine repeating texture of the background.

\paragraph{Mip-NeRF 360}
Figure~\ref{fig:qual_360} extends the comparison to unbounded $360^\circ$ scenes, where the wider viewpoint distribution and the presence of a complex background make the sparse-view problem substantially harder. We include both indoor (\emph{room}, \emph{bonsai}, \emph{counter}) and outdoor (\emph{garden}) cases. Vanilla 3DGS shows two characteristic failures here: large washed-out patches in indoor scenes with strong lighting variation, where Gaussians collapse into low-opacity floaters that cannot be disambiguated from the few training views; and loss of thin foreground structures---the slats and arm rests of the \emph{garden} bench, the stems and petals of the \emph{bonsai}, the utensils on the kitchen \emph{counter}, the picture frames in \emph{room}---against textured backgrounds. Our reconstructions retain sharper object boundaries on these thin structures and preserve the appearance of the background, indicating that the additional pseudo-view supervision compensates for the under-coverage of the training camera distribution rather than merely smoothing the rendered output.

\subsection{Quantitative Results}
\label{sec:quantitative}

We report PSNR, SSIM~\cite{wang2004ssim} and LPIPS~\cite{zhang2018lpips} on the held-out test split (every $8$th image, \texttt{llffhold=8}) following the standard 3DGS evaluation protocol. Per-scene PSNR and average metrics on LLFF are listed in Table~\ref{tab:llff_results}; the corresponding numbers on Mip-NeRF~360 are listed in Table~\ref{tab:mipnerf360_results}.

\begin{table}[!t]
  \centering
  \scriptsize
  \setlength{\tabcolsep}{3pt}
  \caption{Per-scene PSNR and average metrics on LLFF (\texttt{llffhold=8}, \texttt{-r 8}). \emph{Avg} rows summarize PSNR, SSIM and LPIPS over the eight scenes. Best per column in bold.}
  \label{tab:llff_results}
  \begin{tabular}{l cc cc cc}
    \toprule
              & \multicolumn{2}{c}{3-view} & \multicolumn{2}{c}{6-view} & \multicolumn{2}{c}{9-view} \\
    \cmidrule(lr){2-3} \cmidrule(lr){4-5} \cmidrule(lr){6-7}
    Scene     & 3DGS  & Ours           & 3DGS  & Ours           & 3DGS  & Ours           \\
    \midrule
    fern      & 21.21 & \cellcolor{cellgreen1}\textbf{22.64} & 24.03 & \cellcolor{cellgreen1}\textbf{25.15} & 26.32 & \cellcolor{cellgreen2}\textbf{27.25} \\
    flower    & 19.22 & \cellcolor{cellgreen1}\textbf{20.65} & 24.31 & \cellcolor{cellgreen2}\textbf{25.13} & 25.99 & \cellcolor{cellgreen2}\textbf{26.63} \\
    fortress  & \textbf{21.24} & \cellcolor{cellorange}19.35 & 26.10 & \cellcolor{cellgreen1}\textbf{27.33} & 28.37 & \cellcolor{cellgreen1}\textbf{29.49} \\
    horns     & 18.51 & \cellcolor{cellyellow}\textbf{18.61} & 22.72 & \cellcolor{cellgreen1}\textbf{23.73} & 25.43 & \cellcolor{cellgreen2}\textbf{26.25} \\
    leaves    & 16.53 & \cellcolor{cellgreen2}\textbf{17.25} & 18.84 & \cellcolor{cellgreen2}\textbf{19.57} & 20.88 & \cellcolor{cellgreen2}\textbf{21.54} \\
    orchids   & 15.14 & \cellcolor{cellgreen2}\textbf{16.01} & 16.59 & \cellcolor{cellgreen2}\textbf{17.54} & 18.09 & \cellcolor{cellgreen2}\textbf{19.05} \\
    room      & 21.05 & \cellcolor{cellgreen3}\textbf{21.58} & 27.67 & \cellcolor{cellgreen3}\textbf{28.19} & 28.50 & \cellcolor{cellgreen3}\textbf{28.95} \\
    trex      & 20.29 & \cellcolor{cellyellow}\textbf{20.30} & 23.80 & \cellcolor{cellgreen2}\textbf{24.58} & 25.97 & \cellcolor{cellyellow}\textbf{25.98} \\
    \midrule
    PSNR avg  & 19.15 & \cellcolor{cellgreen3}\textbf{19.55} & 23.01 & \cellcolor{cellgreen2}\textbf{23.90} & 24.94 & \cellcolor{cellgreen2}\textbf{25.64} \\
    SSIM avg  & 0.6460 & \cellcolor{cellgreen2}\textbf{0.6848} & 0.7961 & \cellcolor{cellgreen2}\textbf{0.8248} & 0.8531 & \cellcolor{cellgreen3}\textbf{0.8719} \\
    LPIPS avg & 0.2362 & \cellcolor{cellgreen1}\textbf{0.2079} & 0.1406 & \cellcolor{cellgreen2}\textbf{0.1184} & 0.1031 & \cellcolor{cellgreen3}\textbf{0.0899} \\
    \bottomrule
  \end{tabular}
\end{table}

\paragraph{Aggregate improvements}
On LLFF, average PSNR improves over vanilla 3DGS by $+0.40$, $+0.89$ and $+0.70$~dB at 3, 6 and 9 views, with consistent SSIM gains ($+0.0388$, $+0.0287$, $+0.0188$) and LPIPS reductions ($-0.0283$, $-0.0222$, $-0.0132$). On Mip-NeRF~360, average PSNR rises by $+1.18$~dB at 12 views and $+0.80$~dB at 24 views, with corresponding SSIM gains of $+0.0469$ and $+0.0260$, and---unlike the quarter-resolution configuration we previously reported---with LPIPS reductions on both view counts ($-0.0048$ and $-0.0012$).

\paragraph{Per-scene LLFF}
Our method outperforms vanilla 3DGS on $23$ of the $24$ per-scene LLFF cells in Table~\ref{tab:llff_results}; the only regression is \emph{fortress} 3-view, where the diffusion pseudo-views yield a $-1.89$~dB drop (from $21.24$ to $19.35$). Notably, this single per-scene regression also dominates the 3-view average: excluding \emph{fortress}, the per-scene 3-view improvements average $+0.73$~dB---comparable to the 9-view average ($+0.70$~dB)---rather than the overall $+0.40$~dB figure reported in Table~\ref{tab:llff_results}. The largest gains appear on scenes with substantial unobserved volume or thin repeated structure: \emph{flower} 3-view ($+1.44$~dB), \emph{fern} 3-view ($+1.43$~dB), \emph{fortress} 6-view ($+1.23$~dB), \emph{fern} 6-view ($+1.12$~dB), \emph{fortress} 9-view ($+1.12$~dB) and \emph{horns} 6-view ($+1.01$~dB).

\begin{table}[!t]
  \centering
  \small
  \setlength{\tabcolsep}{5pt}
  \caption{Per-scene PSNR and average metrics on Mip-NeRF~360 (\texttt{llffhold=8}, \texttt{-r 8}). Best per column in bold.}
  \label{tab:mipnerf360_results}
  \begin{tabular}{l cc cc}
    \toprule
              & \multicolumn{2}{c}{12-view} & \multicolumn{2}{c}{24-view} \\
    \cmidrule(lr){2-3} \cmidrule(lr){4-5}
    Scene     & 3DGS  & Ours           & 3DGS  & Ours           \\
    \midrule
    bicycle   & 16.37 & \cellcolor{cellgreen1}\textbf{18.05} & 19.63 & \cellcolor{cellgreen1}\textbf{21.04} \\
    bonsai    & 18.85 & \cellcolor{cellgreen2}\textbf{19.44} & 23.55 & \cellcolor{cellgreen2}\textbf{24.53} \\
    counter   & 17.71 & \cellcolor{cellgreen1}\textbf{18.73} & 22.42 & \cellcolor{cellgreen3}\textbf{22.86} \\
    garden    & 19.43 & \cellcolor{cellgreen1}\textbf{20.79} & 24.01 & \cellcolor{cellgreen2}\textbf{24.82} \\
    kitchen   & 18.94 & \cellcolor{cellgreen3}\textbf{19.29} & 22.64 & \cellcolor{cellgreen2}\textbf{23.50} \\
    room      & 19.61 & \cellcolor{cellgreen2}\textbf{20.52} & \textbf{24.11} & \cellcolor{cellorange}23.94 \\
    stump     & 15.96 & \cellcolor{cellgreen1}\textbf{18.27} & 19.25 & \cellcolor{cellgreen1}\textbf{20.51} \\
    \midrule
    PSNR avg  & 18.12 & \cellcolor{cellgreen1}\textbf{19.30} & 22.23 & \cellcolor{cellgreen2}\textbf{23.03} \\
    SSIM avg  & 0.5124 & \cellcolor{cellgreen1}\textbf{0.5593} & 0.6851 & \cellcolor{cellgreen2}\textbf{0.7111} \\
    LPIPS avg & 0.3719 & \cellcolor{cellgreen3}\textbf{0.3671} & 0.2424 & \cellcolor{cellyellow}\textbf{0.2412} \\
    \bottomrule
  \end{tabular}
\end{table}

\paragraph{Per-scene Mip-NeRF~360}
On Mip-NeRF~360 we improve PSNR on $13$ of the $14$ per-scene cells in Table~\ref{tab:mipnerf360_results}, with the largest absolute gains on the unbounded outdoor scenes \emph{stump} ($+2.30$~dB) and \emph{bicycle} ($+1.68$~dB) at 12 views. The only per-scene regression is \emph{room} 24-view, where the bounded indoor geometry is already well-constrained by $24$ training cameras and our method trails the baseline by a negligible $0.17$~dB; bounded indoor scenes as a class show more moderate improvements than the unbounded outdoor class, consistent with the observation that diffusion supervision matters most where the training cameras leave a large unobserved volume.

\paragraph{LPIPS on Mip-NeRF~360}
At \texttt{-r 8}, our method now improves LPIPS on Mip-NeRF~360 as well, reducing the average by $-0.0048$ at 12 views and $-0.0012$ at 24 views. The perceptual gap in the quarter-resolution configuration we previously reported was driven by subtle texture drift on unbounded backgrounds---especially the sky and distant foliage of \emph{stump} and \emph{bicycle}---that is no longer measurable at eighth resolution, where the VGG perceptual loss averages over larger effective receptive fields and the background-texture differences become perceptually negligible.

\subsection{Comparison with Prior Methods}
\label{sec:sota}

Table~\ref{tab:sota} compares our method with a selection of sparse-view NVS methods on LLFF (3-view) and Mip-NeRF~360 (24-view).  We include three NeRF-based methods---RegNeRF~\cite{niemeyer2022regnerf}, FreeNeRF~\cite{yang2023freenerf}, SparseNeRF~\cite{wang2023sparsenerf}---and DNGaussian~\cite{li2024dngaussian}, a Gaussian-based method from CVPR 2024.  LLFF results for NeRF-based methods are taken from~\cite{zhang2024cor_gs} (VGG LPIPS, $\tfrac{1}{8}$ resolution); DNGaussian LLFF results are from its original paper~\cite{li2024dngaussian} (same $\tfrac{1}{8}$ resolution, but without dense MVS initialization).  Mip-NeRF~360 results for NeRF-based methods are from~\cite{zhu2024fsgs} ($\dagger$, may differ in LPIPS backbone); DNGaussian is not evaluated on Mip-NeRF~360 in its original paper.  We note that this comparison is not exhaustive; several concurrent methods address the same problem and are not included here due to protocol or data differences.

\paragraph{LLFF 3-view}
Among the compared methods, our method obtains higher SSIM ($0.685$) and lower LPIPS ($0.208$) than all four baselines.  In PSNR, SparseNeRF ($19.86$) and FreeNeRF ($19.63$) outperform our method ($19.55$); we attribute this to their depth-ranking supervision providing stronger per-pixel accuracy on forward-facing scenes, whereas our gains are more pronounced on perceptual metrics.

\paragraph{Mip-NeRF 360 24-view}
On the unbounded benchmark, our method obtains higher PSNR ($23.03$), SSIM ($0.711$), and lower LPIPS ($0.241$) than all compared methods.  The gap to the next best PSNR entry (SparseNeRF: $22.85$) is $+0.18$~dB, consistent with the observation in Section~\ref{sec:discussion} that diffusion-supervised pseudo-views contribute most in unbounded $360^\circ$ scenes where the unobserved volume is large.

\begin{table*}[!t]
  \centering
  \small
  \setlength{\tabcolsep}{7pt}
  \caption{Comparison with sparse-view NVS methods on LLFF (3-view) and Mip-NeRF~360 (24-view). NeRF-based LLFF results are from~\cite{zhang2024cor_gs} (VGG LPIPS, $\tfrac{1}{8}$ resolution); DNGaussian LLFF results are from~\cite{li2024dngaussian} (same $\tfrac{1}{8}$ resolution). $\dagger$~Mip-NeRF~360 results for NeRF-based methods are from~\cite{zhu2024fsgs}. \textbf{Bold}: best; \underline{underline}: second best (per column, excluding baseline).}
  \label{tab:sota}
  \begin{tabular}{l ccc ccc}
    \toprule
    \multirow{2}{*}{Method}
      & \multicolumn{3}{c}{LLFF (3-view)}
      & \multicolumn{3}{c}{Mip-NeRF~360 (24-view)} \\
    \cmidrule(lr){2-4}\cmidrule(lr){5-7}
    & PSNR$\uparrow$ & SSIM$\uparrow$ & LPIPS$\downarrow$
      & PSNR$\uparrow$ & SSIM$\uparrow$ & LPIPS$\downarrow$ \\
    \midrule
    RegNeRF~\cite{niemeyer2022regnerf}
      & 19.08 & 0.587 & 0.336
      & $22.19^\dagger$ & $0.643^\dagger$ & $0.335^\dagger$ \\
    FreeNeRF~\cite{yang2023freenerf}
      & \underline{19.63} & 0.612 & 0.308
      & $22.78^\dagger$ & $0.689^\dagger$ & $0.323^\dagger$ \\
    SparseNeRF~\cite{wang2023sparsenerf}
      & \textbf{19.86} & 0.624 & 0.328
      & $\underline{22.85}^\dagger$ & $\underline{0.693}^\dagger$ & $\underline{0.315}^\dagger$ \\
    DNGaussian~\cite{li2024dngaussian}
      & 19.12 & 0.591 & \underline{0.294}
      & --- & --- & --- \\
    \midrule
    Vanilla 3DGS~\cite{kerbl20233dgs}
      & 19.15 & 0.646 & 0.236
      & 22.23 & 0.685 & 0.242 \\
    \textbf{GS-GS (ours)}
      & 19.55 & \textbf{0.685} & \textbf{0.208}
      & \textbf{23.03} & \textbf{0.711} & \textbf{0.241} \\
    \bottomrule
  \end{tabular}
\end{table*}

\subsection{Ablation Study}
\label{sec:ablation}

To isolate the contribution of the two main components introduced by our method---generative active pseudo-view selection (GAPS, Section~\ref{sec:gaps}) and density-adaptive DropGaussian regularization (Section~\ref{sec:dropgaussian})---we perform a component-wise ablation on the Mip-NeRF~360 dataset in the 24-view configuration. We choose this setting because it is sparse enough for the diffusion supervision to make a measurable difference, while the per-scene statistics are stable enough across the seven scenes for the gap between configurations to be interpretable. All variants share the same sparse MVS initialization, the same SDXL/ControlNet/IP-Adapter/LoRA stack, the same training budget of $10{,}000$ iterations, and the same 3DGS optimizer; only the ablated component is changed.

\paragraph{Ablation conditions}
We compare four variants. \textbf{Vanilla 3DGS} omits the diffusion supervision entirely and corresponds to the baseline reported in Table~\ref{tab:mipnerf360_results}. \textbf{Ours w/o GAPS} replaces the GAPS scoring function, the dual-criterion admission gate, and the per-phase refresh with a minimal static baseline: a deterministic set of eight pair-midpoint pseudo-views ($\alpha{=}0.5$, interpolation only) generated once in the first diffusion phase and never refreshed; the rest of the constrained generation stack (SDXL, ControlNet, IP-Adapter, LoRA) and DropGaussian remain identical to the full method. \textbf{Ours w/o DropGaussian} removes the density-adaptive Gaussian dropout regularizer and keeps everything else identical to the full method. \textbf{Ours (full)} is the configuration reported in Section~\ref{sec:quantitative}.

\begin{table}[!t]
  \centering
  \small
  \setlength{\tabcolsep}{4pt}
  \caption{Component ablation on Mip-NeRF~360 (24-view, \texttt{llffhold=8}, \texttt{-r 8}). \emph{w/o GAPS} replaces the scored active selection and per-phase refresh with a static set of pair-midpoint pseudo-views (see text); \emph{w/o DropGaussian} removes the density-adaptive dropout regularizer. Best per column in bold.}
  \label{tab:ablation_360}
  \begin{tabular}{l ccc}
    \toprule
    Variant                  & PSNR           & SSIM            & LPIPS \\
    \midrule
    Vanilla 3DGS             & 22.23          & 0.6851          & 0.2424 \\
    Ours w/o DropGaussian    & 22.62          & 0.6923          & 0.2453 \\
    Ours w/o GAPS            & 22.88          & 0.7002          & 0.2497 \\
    \midrule
    Ours (full)              & \textbf{23.03} & \textbf{0.7111} & \textbf{0.2412} \\
    \bottomrule
  \end{tabular}
\end{table}

\paragraph{Component contributions}
Table~\ref{tab:ablation_360} reports the average PSNR, SSIM and LPIPS over the seven Mip-NeRF~360 scenes. Removing DropGaussian drops PSNR from $23.03$ to $22.62$~dB ($-0.41$~dB), SSIM from $0.7111$ to $0.6923$ ($-0.0188$), and inflates LPIPS by $+0.0041$, confirming that the density-adaptive regularizer supplies the largest single-component slice of the $+0.80$~dB gap between vanilla 3DGS and our full method. Removing GAPS produces a more modest PSNR penalty ($-0.15$~dB, $23.03 \rightarrow 22.88$) and a comparable SSIM drop ($-0.0109$, $0.7111 \rightarrow 0.7002$), but---more notably---inflates LPIPS to $0.2497$, which is \emph{worse} than the vanilla baseline ($0.2424$) that uses no pseudo-views at all. In other words, indiscriminate pseudo-view supervision injects perceptual artifacts that the vanilla optimization never sees; the scored active selection and the per-phase refresh are precisely what keep the hallucinated supervision perceptually clean. SSIM still improves over vanilla in this ablated variant ($0.7002$ vs $0.6851$), so the static pseudo-views do help resolve structural ambiguity in unobserved regions even without scoring---they just do so at a perceptual cost. Both ablated variants still outperform vanilla 3DGS in PSNR and SSIM, indicating that each component is individually beneficial; LPIPS is the axis on which GAPS matters most under this setting.

\paragraph{DropGaussian and the sparse-view regime}
The $-0.41$~dB PSNR penalty from removing DropGaussian is the largest single-component penalty in Table~\ref{tab:ablation_360} and is comparable in magnitude to the per-view-count improvements our method delivers on LLFF (Section~\ref{sec:quantitative}), suggesting that the density-adaptive dropout is acting as a global anti-overfitting mechanism for the Gaussian distribution whose contribution is largely independent of dataset geometry (forward-facing LLFF vs.\ unbounded $360^\circ$). The trigger only enables dropout when the local Gaussian density exceeds the threshold $\rho > 2 \times 10^5$, so scenes that never densify past this threshold are unaffected, which is why the per-scene penalty is concentrated on the bounded indoor scenes (\emph{bonsai}, \emph{kitchen}) where the Gaussian count grows most aggressively. Removing DropGaussian also inflates LPIPS ($+0.0041$), which we attribute to the same over-fitting mode: an un-regularized Gaussian distribution places density in regions where the reprojection residual of the few training views is lowest, producing images that look plausible on the training frames but drift on the held-out views.

\subsection{Discussion}
\label{sec:discussion}

\paragraph{Where the gains come from}
The quantitative results in Section~\ref{sec:quantitative} share a single trend across both datasets: the absolute improvement over vanilla 3DGS scales with the size of the unobserved volume that the training cameras leave behind. The largest gains appear in the most under-constrained settings---Mip-NeRF~360 12-view ($+1.18$~dB on average, $+2.30$~dB on \emph{stump} and $+1.68$~dB on \emph{bicycle}) and LLFF 6-view ($+0.89$~dB on average)---and then moderate in either direction as more or fewer views are added. LLFF 3-view is a partial exception ($+0.40$~dB on average), but as discussed under \emph{Per-scene LLFF} above this is driven almost entirely by the \emph{fortress} 3-view regression: on the other seven scenes the per-scene mean is $+0.73$~dB, comparable to the 9-view setting, so the 3-view regime is not systematically harder for our method but rather exposes the sensitivity of diffusion supervision to the specific geometry of a single outlier scene. This is exactly the regime in which the two design choices of Section~\ref{sec:method} are most useful: the GAPS scoring function (Section~\ref{sec:gaps}) places pseudo-views in the unobserved regions where the optimizer has no gradient signal, and the four-fold constrained generation pipeline (ControlNet, IP-Adapter, LoRA, img2img) keeps those samples geometrically consistent enough to be used as supervision rather than as noise. The qualitative renders in Figures~\ref{fig:qual_llff} and~\ref{fig:qual_360} corroborate the same picture from the appearance side: the scenes with the largest PSNR gains are precisely those dominated by thin or repeated structure---bone, fronds, slats, stems---where vanilla 3DGS collapses into floaters and our reconstructions retain sharp boundaries.

\paragraph{Component complementarity}
The ablation in Table~\ref{tab:ablation_360} shows that DropGaussian and GAPS act on different failure modes and are complementary rather than redundant: removing DropGaussian costs $-0.41$~dB of PSNR with a small LPIPS penalty ($+0.0041$), whereas removing GAPS costs a more modest $-0.15$~dB of PSNR but inflates LPIPS by $+0.0085$ to a value ($0.2497$) that is \emph{worse} than the vanilla baseline with no pseudo-views at all. This split is consistent with the role we ascribe to each component in Section~\ref{sec:method}. DropGaussian acts on the \emph{geometric} failure mode of sparse-view 3DGS---unregularized Gaussian proliferation that over-fits the few training frames---and its contribution is largely independent of how pseudo-views are chosen, which is why its penalty shows up primarily in PSNR and SSIM. GAPS and its dual-criterion admission gate act on the \emph{appearance} failure mode of pseudo-view supervision: once the pool is replaced by static pair-midpoint samples, the optimizer is forced to absorb whatever the diffusion model hallucinates at those fixed camera poses, and the resulting renders drift perceptually even as structural fidelity (SSIM) still improves over vanilla. The LPIPS regression below vanilla shows that indiscriminate pseudo-view supervision is not merely sub-optimal but actively harmful on the perceptual axis, and that the scored selection together with the per-phase refresh is what keeps the hallucinated supervision perceptually safe. With $24$ training cameras around an unbounded $360^\circ$ scene, most of the surface is already well-represented and the PSNR headroom is small; in the more under-constrained LLFF 3- and 6-view settings of Table~\ref{tab:llff_results}, where average gains are $+0.40$~dB and $+0.89$~dB, we expect the PSNR contribution of GAPS to grow proportionally, though isolating this across view counts would require a matched ablation on LLFF that we leave to future work.

\paragraph{The Mip-NeRF~360 LPIPS gap}
Our method improves the average LPIPS on both benchmarks and every evaluated view count at the \texttt{-r 8} protocol reported in this paper: LLFF records reductions of $-0.0283$, $-0.0222$ and $-0.0132$ at $3/6/9$ views, and Mip-NeRF~360 records $-0.0048$ at $12$ views and $-0.0012$ at $24$ views (Tables~\ref{tab:llff_results}, \ref{tab:mipnerf360_results}). The Mip-NeRF~360 reductions are an order of magnitude smaller than the LLFF reductions, which we attribute to the dominant role the unbounded background plays in the $360^\circ$ LPIPS: sky and distant foliage account for most of the pixel mass in the held-out views, and at eighth resolution the VGG receptive fields average over large enough patches that subtle texture drift in those regions has a negligible effect on the perceptual score. The ablation in Table~\ref{tab:ablation_360} is consistent with this reading: both ablated variants degrade LPIPS on Mip-NeRF~360, and removing GAPS is the larger of the two penalties ($+0.0085$ vs $+0.0041$ when removing DropGaussian), pushing LPIPS past the vanilla baseline. The full method is the only configuration that improves LPIPS over vanilla, and it does so precisely because the GAPS scoring, admission gate, and per-phase refresh---in combination with the density-adaptive regularizer---keep the diffusion-supervised background reconstructions perceptually close to the held-out photographs on unbounded scenes.

\paragraph{Practical implications}
Two practical points follow from the experiments. \emph{(i)} The improvements are obtained without any architectural change to the 3DGS optimizer or any modification to the rasterizer, which means the method can be dropped into an existing 3DGS pipeline as an additional alternating training stage. \emph{(ii)} All experiments run on a single $24$~GB consumer GPU within the same $10{,}000$-iteration training budget as vanilla 3DGS, so the gains do not come from a longer optimization or from a heavier rendering stack; the cost is the diffusion forward pass at the phase boundaries, which is amortized across the rest of the training schedule.

\section{Conclusion}
\label{sec:conclusion}
We presented an alternating optimization framework that augments sparse-view 3D Gaussian Splatting with diffusion-generated pseudo-views. The framework rests on three design choices that target distinct failure modes of the under-constrained sparse-view problem. First, we constrain a pre-trained image diffusion model with a four-fold pipeline---depth-conditioned ControlNet, IP-Adapter style transfer, scene-specific LoRA fine-tuning, and img2img structural anchoring---so that every pseudo-view is geometrically and stylistically consistent with the reconstructed scene rather than being a free generative sample. Second, we introduce Generative Active Pseudo-view Selection (GAPS), which scores candidate viewpoints by jointly balancing reconstruction informativeness and generative reliability and follows an annealing schedule that progresses from conservative interpolation in early training to exploratory extrapolation in later stages. A dual-criterion admission gate together with per-pixel uncertainty-weighted losses prevents unreliable generations from destabilizing training. Third, we add a density-adaptive DropGaussian regularizer that activates only where the local Gaussian density exceeds a threshold, providing a global anti-overfitting mechanism that is largely orthogonal to scene geometry.

Experiments on LLFF and Mip-NeRF~360 show consistent improvements over vanilla 3DGS across all view counts. On LLFF, average PSNR rises by $+0.40$, $+0.89$ and $+0.70$~dB at 3, 6 and 9 views, with simultaneous SSIM gains ($+0.0388$, $+0.0287$, $+0.0188$) and LPIPS reductions ($-0.0283$, $-0.0222$, $-0.0132$); on Mip-NeRF~360, average PSNR rises by $+1.18$~dB at 12 views and $+0.80$~dB at 24 views, with SSIM gains of $+0.0469$ and $+0.0260$ and LPIPS reductions of $-0.0048$ and $-0.0012$. The largest absolute improvements appear in the most under-constrained settings and on scenes dominated by thin or repeated structure, in line with the central design hypothesis that diffusion supervision is most useful where the training cameras leave the largest unobserved volume to fill. Component-wise ablation on Mip-NeRF~360 confirms that GAPS and density-adaptive DropGaussian act on complementary failure modes: removing DropGaussian produces the largest PSNR penalty ($-0.41$~dB), whereas removing GAPS inflates LPIPS to $0.2497$---worse than the vanilla baseline ($0.2424$) with no pseudo-views---showing that indiscriminate pseudo-view supervision is actively harmful on the perceptual axis and that the scored selection together with the per-phase refresh are what keep the hallucinated supervision perceptually clean. The full method is the only configuration that simultaneously improves photometric, structural, and perceptual metrics on every benchmark.

Several directions remain open. The Mip-NeRF~360 LPIPS reductions, while consistently positive, remain an order of magnitude smaller than the LLFF reductions at the same eighth resolution, indicating that the unbounded background---where sky and distant foliage dominate the pixel mass---still leaves residual perceptual headroom that a domain-specific photographic prior or a tighter IP-Adapter style constraint on background regions could plausibly close. Extending the alternating framework beyond static scenes to dynamic or relightable settings, and exploring video diffusion priors to enforce temporal consistency across pseudo-view batches, are natural next steps. More broadly, our results indicate that pre-trained 2D generative priors can supply the missing supervision for under-constrained 3D reconstruction problems when their outputs are gated and weighted carefully---a principle that extends well beyond the specific 3DGS pipeline studied here.

\section*{Acknowledgements}
This work was supported by the Goertek. The authors would like to thank the members of the Alpha Labs for their valuable discussions and technical support during the research.

\section*{CRediT authorship contribution statement}
\textbf{Hongfei Zhu:} Writing -- original draft, Visualization, Validation, Software, Methodology, Investigation, Formal analysis, Data curation, Conceptualization.
\textbf{Haochen Deng:} Writing -- review \& editing, Validation.
\textbf{Sitao Zhang:} Writing -- review \& editing, Resources.
\textbf{Ling Zhou:} Writing -- review \& editing, Supervision, Conceptualization.

\section*{Declaration of competing interest}
The authors declare that they have no known competing financial interests or personal relationships that could have appeared to influence the work reported in this paper.

\section*{Data availability}
Data will be made available on request.

\section*{Declaration of generative AI and AI-assisted technologies in the writing process}
During the preparation of this work the authors used Claude (Anthropic) in order to polish language and improve the readability of the manuscript. After using this tool, the authors reviewed and edited the content as needed and take full responsibility for the content of the publication.


\bibliographystyle{elsarticle-num}
\renewcommand\bibfont{\small}
\setlength{\bibsep}{0pt plus 0.3ex}
\bibliography{references}

\begin{thebibliography}{10}
\expandafter\ifx\csname url\endcsname\relax
  \def\url#1{\texttt{#1}}\fi
\expandafter\ifx\csname urlprefix\endcsname\relax\def\urlprefix{URL }\fi
\expandafter\ifx\csname href\endcsname\relax
  \def\href#1#2{#2} \def\path#1{#1}\fi

\bibitem{kerbl20233dgs}
B.~Kerbl, G.~Kopanas, T.~Leimk{\"u}hler, G.~Drettakis, 3d gaussian splatting
  for real-time radiance field rendering, ACM Transactions on Graphics 42~(4)
  (2023) 139:1--139:14.
\newblock \href {https://doi.org/10.1145/3592433} {\path{doi:10.1145/3592433}}.

\bibitem{li2024dngaussian}
J.~Li, J.~Zhang, X.~Bai, J.~Zheng, X.~Ning, J.~Zhou, L.~Gu, {DNGaussian}:
  Optimizing sparse-view {3D} gaussian radiance fields with global-local depth
  normalization, in: Proceedings of the IEEE/CVF Conference on Computer Vision
  and Pattern Recognition (CVPR), 2024, pp. 20775--20785.
\newblock \href {https://doi.org/10.1109/CVPR52733.2024.01963}
  {\path{doi:10.1109/CVPR52733.2024.01963}}.

\bibitem{zhang2024cor_gs}
J.~Zhang, J.~Li, X.~Yu, L.~Huang, L.~Gu, J.~Zheng, X.~Bai, {CoR-GS}:
  Sparse-view {3D Gaussian Splatting} via co-regularization, in: Proceedings of
  the European Conference on Computer Vision (ECCV), 2024, pp. 335--352.
\newblock \href {https://doi.org/10.1007/978-3-031-73232-4_19}
  {\path{doi:10.1007/978-3-031-73232-4_19}}.

\bibitem{park2025dropgaussian}
H.~Park, G.~Ryu, W.~Kim, {DropGaussian}: Structural regularization for
  sparse-view {Gaussian} splatting, in: Proceedings of the IEEE/CVF Conference
  on Computer Vision and Pattern Recognition (CVPR), 2025, pp. 21600--21609.
\newblock \href {https://doi.org/10.1109/CVPR52734.2025.02012}
  {\path{doi:10.1109/CVPR52734.2025.02012}}.

\bibitem{kong2025gsgs}
H.~Kong, X.~Yang, X.~Wang, Generative sparse-view {Gaussian} splatting, in:
  Proceedings of the IEEE/CVF Conference on Computer Vision and Pattern
  Recognition (CVPR), 2025, pp. 26745--26755.
\newblock \href {https://doi.org/10.1109/CVPR52734.2025.02491}
  {\path{doi:10.1109/CVPR52734.2025.02491}}.

\bibitem{topaloglu2025oraclegs}
A.~Topalo\u{g}lu, K.~Li, M.~Niemeyer, N.~Navab, A.~M. Tekalp, F.~Tombari,
  {OracleGS}: Grounding generative priors for sparse-view {Gaussian} splatting,
  arXiv preprint arXiv:2509.23258 (2025).

\bibitem{zhang2023controlnet}
L.~Zhang, A.~Rao, M.~Agrawala, Adding conditional control to text-to-image
  diffusion models, in: Proceedings of the IEEE/CVF International Conference on
  Computer Vision (ICCV), 2023, pp. 3836--3847.
\newblock \href {https://doi.org/10.1109/ICCV51070.2023.00355}
  {\path{doi:10.1109/ICCV51070.2023.00355}}.

\bibitem{ye2023ipadapter}
H.~Ye, J.~Zhang, S.~Liu, X.~Han, W.~Yang, {IP-Adapter}: Text compatible image
  prompt adapter for text-to-image diffusion models, arXiv preprint
  arXiv:2308.06721 (2023).
\newblock \href {https://doi.org/10.48550/arXiv.2308.06721}
  {\path{doi:10.48550/arXiv.2308.06721}}.

\bibitem{hu2022lora}
E.~J. Hu, Y.~Shen, P.~Wallis, Z.~Allen-Zhu, Y.~Li, S.~Wang, L.~Wang, W.~Chen,
  \href{https://openreview.net/forum?id=nZeVKeeFYf9}{{LoRA}: Low-rank
  adaptation of large language models}, in: International Conference on
  Learning Representations (ICLR), 2022.
\newline\urlprefix\url{https://openreview.net/forum?id=nZeVKeeFYf9}

\bibitem{mildenhall2020nerf}
B.~Mildenhall, P.~P. Srinivasan, M.~Tancik, J.~T. Barron, R.~Ramamoorthi,
  R.~Ng, {NeRF}: Representing scenes as neural radiance fields for view
  synthesis, in: Proceedings of the European Conference on Computer Vision
  (ECCV), 2020, pp. 405--421.
\newblock \href {https://doi.org/10.1007/978-3-030-58452-8_24}
  {\path{doi:10.1007/978-3-030-58452-8_24}}.

\bibitem{barron2022mipnerf360}
J.~T. Barron, B.~Mildenhall, D.~Verbin, P.~P. Srinivasan, P.~Hedman, {Mip-NeRF}
  360: Unbounded anti-aliased neural radiance fields, in: Proceedings of the
  IEEE/CVF Conference on Computer Vision and Pattern Recognition (CVPR), 2022,
  pp. 5470--5479.
\newblock \href {https://doi.org/10.1109/CVPR52688.2022.00539}
  {\path{doi:10.1109/CVPR52688.2022.00539}}.

\bibitem{jain2021dietnerf}
A.~Jain, M.~Tancik, P.~Abbeel, Putting {NeRF} on a diet: Semantically
  consistent few-shot view synthesis, in: Proceedings of the IEEE/CVF
  International Conference on Computer Vision (ICCV), 2021, pp. 5885--5894.
\newblock \href {https://doi.org/10.1109/ICCV48922.2021.00583}
  {\path{doi:10.1109/ICCV48922.2021.00583}}.

\bibitem{niemeyer2022regnerf}
M.~Niemeyer, J.~T. Barron, B.~Mildenhall, M.~S.~M. Sajjadi, A.~Geiger,
  N.~Radwan, {RegNeRF}: Regularizing neural radiance fields for view synthesis
  from sparse inputs, in: Proceedings of the IEEE/CVF Conference on Computer
  Vision and Pattern Recognition (CVPR), 2022, pp. 5480--5490.
\newblock \href {https://doi.org/10.1109/CVPR52688.2022.00540}
  {\path{doi:10.1109/CVPR52688.2022.00540}}.

\bibitem{yang2023freenerf}
J.~Yang, M.~Pavone, Y.~Wang, {FreeNeRF}: Improving few-shot neural rendering
  with free frequency regularization, in: Proceedings of the IEEE/CVF
  Conference on Computer Vision and Pattern Recognition (CVPR), 2023, pp.
  8254--8263.
\newblock \href {https://doi.org/10.1109/CVPR52729.2023.00798}
  {\path{doi:10.1109/CVPR52729.2023.00798}}.

\bibitem{wang2023sparsenerf}
G.~Wang, Z.~Chen, C.~C. Loy, Z.~Liu, {SparseNeRF}: Distilling depth ranking for
  few-shot novel view synthesis, in: Proceedings of the IEEE/CVF International
  Conference on Computer Vision (ICCV), 2023, pp. 9065--9076.
\newblock \href {https://doi.org/10.1109/ICCV51070.2023.00832}
  {\path{doi:10.1109/ICCV51070.2023.00832}}.

\bibitem{zhu2024fsgs}
Z.~Zhu, Z.~Fan, Y.~Jiang, Z.~Wang, {FSGS}: Real-time few-shot view synthesis
  using {Gaussian} splatting, in: Proceedings of the European Conference on
  Computer Vision (ECCV), 2024, pp. 145--163.
\newblock \href {https://doi.org/10.1007/978-3-031-72933-1_9}
  {\path{doi:10.1007/978-3-031-72933-1_9}}.

\bibitem{rombach2022ldm}
R.~Rombach, A.~Blattmann, D.~Lorenz, P.~Esser, B.~Ommer, High-resolution image
  synthesis with latent diffusion models, in: Proceedings of the IEEE/CVF
  Conference on Computer Vision and Pattern Recognition (CVPR), 2022, pp.
  10684--10695.
\newblock \href {https://doi.org/10.1109/CVPR52688.2022.01042}
  {\path{doi:10.1109/CVPR52688.2022.01042}}.

\bibitem{podell2024sdxl}
D.~Podell, Z.~English, K.~Lacey, A.~Blattmann, T.~Dockhorn, J.~M{\"u}ller,
  J.~Penna, R.~Rombach,
  \href{https://openreview.net/forum?id=di52zR8xgf}{{SDXL}: Improving latent
  diffusion models for high-resolution image synthesis}, in: The Twelfth
  International Conference on Learning Representations (ICLR), 2024.
\newline\urlprefix\url{https://openreview.net/forum?id=di52zR8xgf}

\bibitem{poole2023dreamfusion}
B.~Poole, A.~Jain, J.~T. Barron, B.~Mildenhall,
  \href{https://openreview.net/forum?id=FjNys5c7VyY}{{DreamFusion}:
  Text-to-{3D} using {2D} diffusion}, in: The Eleventh International Conference
  on Learning Representations (ICLR), 2023.
\newline\urlprefix\url{https://openreview.net/forum?id=FjNys5c7VyY}

\bibitem{bose2025uar_scenes}
S.~Bose, A.~Dutta, S.~Nag, J.~Zhang, J.~Li, K.~Karydis, A.~K. Roy-Chowdhury,
  Uncertainty-aware diffusion-guided refinement of {3D} scenes, in: Proceedings
  of the IEEE/CVF International Conference on Computer Vision (ICCV), 2025, pp.
  28271--28281.

\bibitem{paul2025gaussian_scenes}
S.~Paul, P.~Kaushik, A.~Yuille,
  \href{https://openreview.net/forum?id=yp1CYo6R0r}{Gaussian scenes: Pose-free
  sparse-view scene reconstruction using depth-enhanced diffusion priors},
  Transactions on Machine Learning Research (2025).
\newline\urlprefix\url{https://openreview.net/forum?id=yp1CYo6R0r}

\bibitem{jiang2024fisherrf}
W.~Jiang, B.~Lei, K.~Daniilidis, {FisherRF}: Active view selection and mapping
  with radiance fields using {Fisher} information, in: Proceedings of the
  European Conference on Computer Vision (ECCV), 2024, pp. 422--440.
\newblock \href {https://doi.org/10.1007/978-3-031-72624-8_24}
  {\path{doi:10.1007/978-3-031-72624-8_24}}.

\bibitem{li2024frequency_view}
M.~M. Li, P.-Y. Lajoie, G.~Beltrame, Frequency-based view selection in
  {Gaussian} splatting reconstruction, arXiv preprint arXiv:2409.16470 (2024).
\newblock \href {https://doi.org/10.48550/arXiv.2409.16470}
  {\path{doi:10.48550/arXiv.2409.16470}}.

\bibitem{kim2024uncertainty_4dgs}
M.~Kim, J.~Lim, B.~Han, {4D Gaussian Splatting} in the wild with
  uncertainty-aware regularization, in: Advances in Neural Information
  Processing Systems (NeurIPS), Vol.~37, 2024.

\bibitem{galappaththige2026predictive_uncertainty}
C.~J. Galappaththige, T.~Gottwald, P.~Stehr, E.~Heinert, N.~Suenderhauf,
  D.~Miller, M.~Rottmann, Predictive photometric uncertainty in {Gaussian}
  splatting for novel view synthesis, arXiv preprint arXiv:2603.22786 (2026).
\newblock \href {https://doi.org/10.48550/arXiv.2603.22786}
  {\path{doi:10.48550/arXiv.2603.22786}}.

\bibitem{srivastava2014dropout}
N.~Srivastava, G.~Hinton, A.~Krizhevsky, I.~Sutskever, R.~Salakhutdinov,
  Dropout: A simple way to prevent neural networks from overfitting, Journal of
  Machine Learning Research 15~(56) (2014) 1929--1958.

\bibitem{schoenberger2016sfm}
J.~L. Sch\"{o}nberger, J.-M. Frahm, Structure-from-motion revisited, in:
  Proceedings of the IEEE/CVF Conference on Computer Vision and Pattern
  Recognition (CVPR), 2016, pp. 4104--4113.
\newblock \href {https://doi.org/10.1109/CVPR.2016.445}
  {\path{doi:10.1109/CVPR.2016.445}}.

\bibitem{yang2024depthanythingv2}
L.~Yang, B.~Kang, Z.~Huang, Z.~Zhao, X.~Xu, J.~Feng, H.~Zhao, Depth anything
  {V2}, in: Advances in Neural Information Processing Systems (NeurIPS),
  Vol.~37, 2024.

\bibitem{mildenhall2019llff}
B.~Mildenhall, P.~P. Srinivasan, R.~Ortiz-Cayon, N.~K. Kalantari,
  R.~Ramamoorthi, R.~Ng, A.~Kar, Local light field fusion: Practical view
  synthesis with prescriptive sampling guidelines, ACM Transactions on Graphics
  38~(4) (2019) 29:1--29:14.
\newblock \href {https://doi.org/10.1145/3306346.3322980}
  {\path{doi:10.1145/3306346.3322980}}.

\bibitem{wang2004ssim}
Z.~Wang, A.~C. Bovik, H.~R. Sheikh, E.~P. Simoncelli, Image quality assessment:
  From error visibility to structural similarity, IEEE Transactions on Image
  Processing 13~(4) (2004) 600--612.
\newblock \href {https://doi.org/10.1109/TIP.2003.819861}
  {\path{doi:10.1109/TIP.2003.819861}}.

\bibitem{zhang2018lpips}
R.~Zhang, P.~Isola, A.~A. Efros, E.~Shechtman, O.~Wang, The unreasonable
  effectiveness of deep features as a perceptual metric, in: Proceedings of the
  IEEE/CVF Conference on Computer Vision and Pattern Recognition (CVPR), 2018,
  pp. 586--595.
\newblock \href {https://doi.org/10.1109/CVPR.2018.00068}
  {\path{doi:10.1109/CVPR.2018.00068}}.

\end{thebibliography}

\section*{Author biographies}
\small

\noindent
\begin{minipage}[t]{0.28\linewidth}
  \vspace{0pt}
  \includegraphics[width=\linewidth]{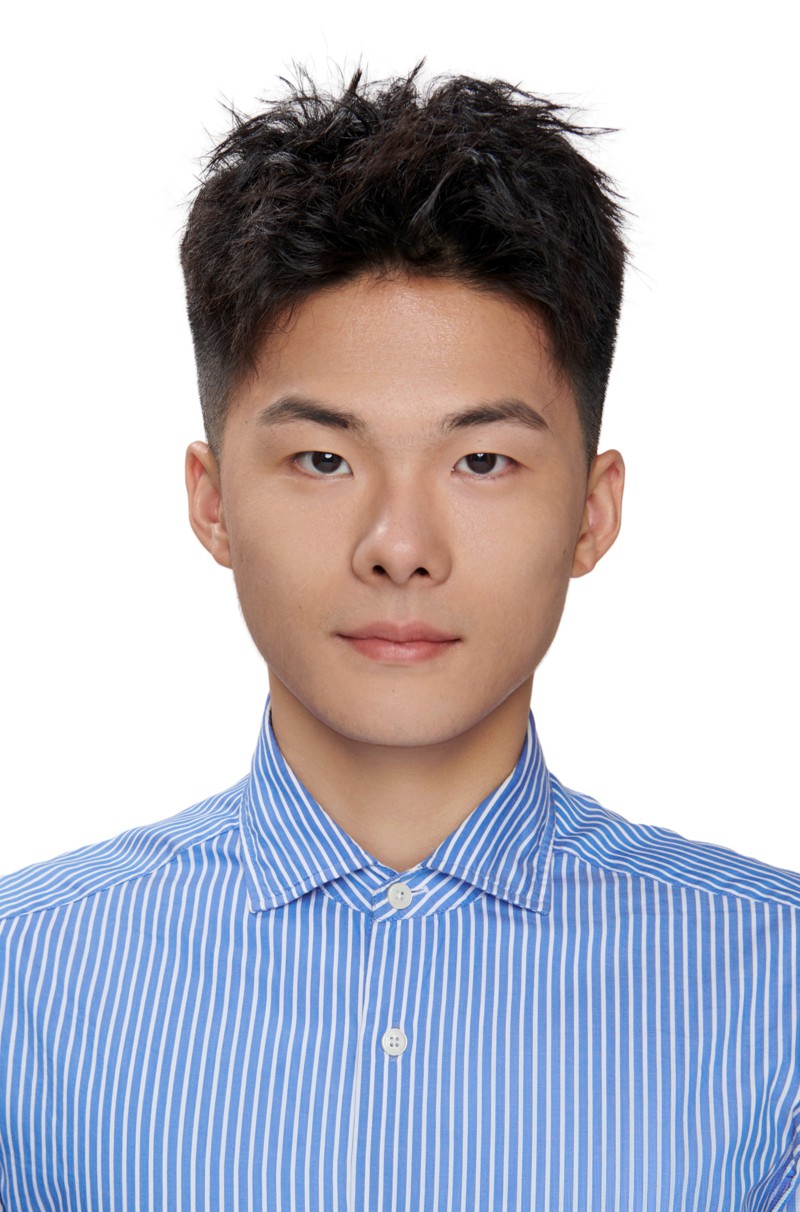}
\end{minipage}\hfill
\begin{minipage}[t]{0.68\linewidth}
  \vspace{0pt}
  \textbf{Hongfei Zhu} is currently a third-year undergraduate student pursuing a B.Eng. in Electronic and Computer Engineering at Shanghai Jiao Tong University, Shanghai, China. He is also a research intern at the Alpha Labs of Goertek Inc., where he works on sparse-view 3D reconstruction and the integration of generative priors into 3D Gaussian Splatting pipelines. His research interests include 3D Gaussian Splatting, simultaneous localization and mapping (SLAM), virtual and augmented reality, and generative models for computer vision.
\end{minipage}

\vspace{1.2em}

\noindent
\begin{minipage}[t]{0.28\linewidth}
  \vspace{0pt}
  \includegraphics[width=\linewidth]{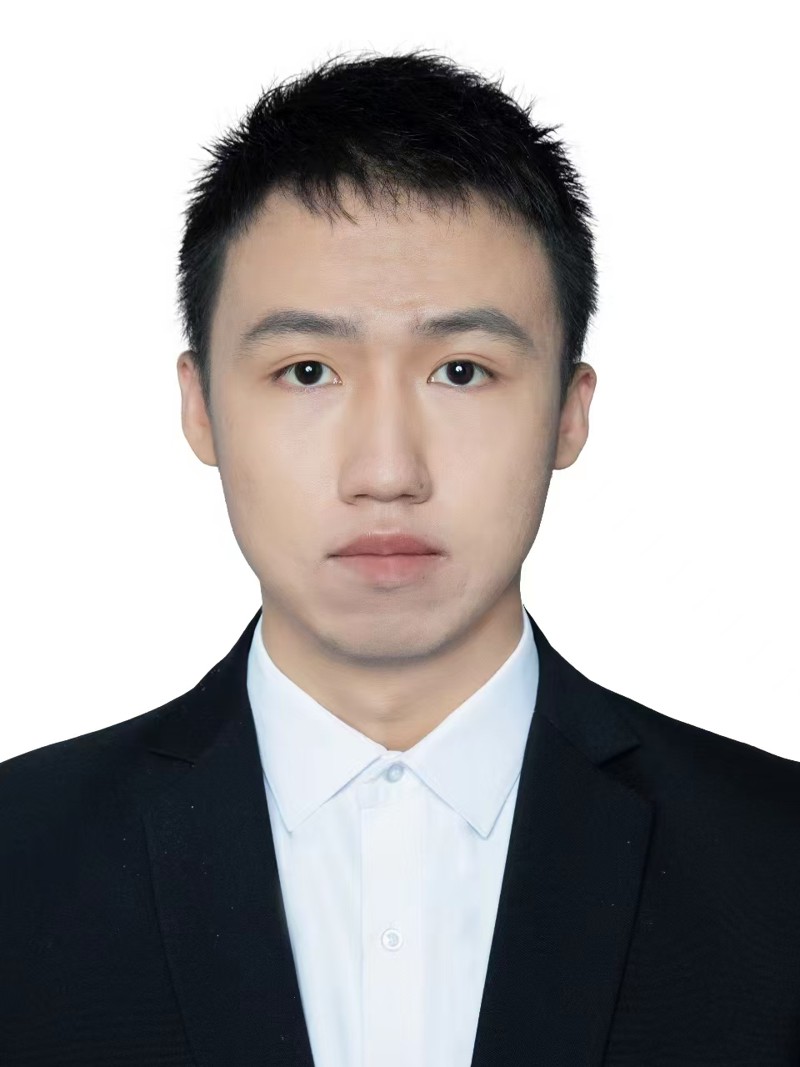}
\end{minipage}\hfill
\begin{minipage}[t]{0.68\linewidth}
  \vspace{0pt}
  \textbf{Haochen Deng} received his B.S. degree from Nanjing University in 2024 and is currently pursuing his M.S. degree at Fudan University, Shanghai, China. His research interests focus on 4D Gaussian Splatting, including efficient dynamic scene representation, reconstruction, and real-time novel view synthesis.
\end{minipage}

\vspace{1.2em}

\noindent
\begin{minipage}[t]{0.28\linewidth}
  \vspace{0pt}
  \includegraphics[width=\linewidth]{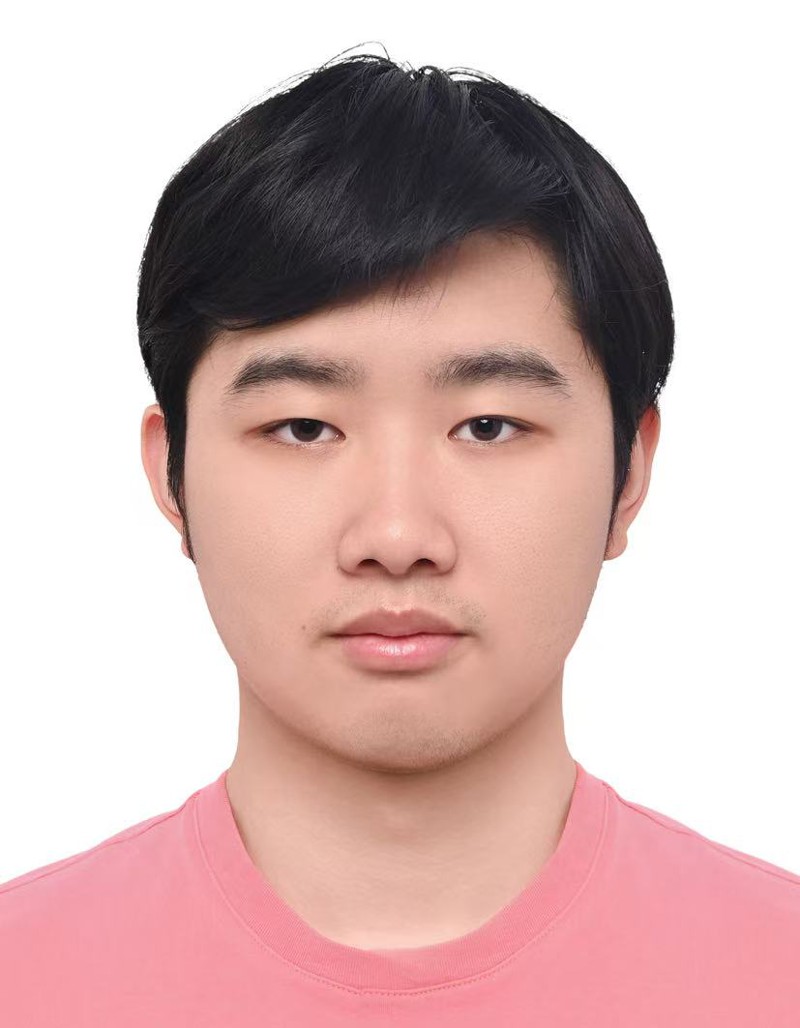}
\end{minipage}\hfill
\begin{minipage}[t]{0.68\linewidth}
  \vspace{0pt}
  \textbf{Sitao Zhang} is currently a Master's candidate in Information Technology and Management at the University of Sydney, Australia. He received his B.Sc. in Mathematics from the East China University of Science and Technology. He is also a member of the XR engineering team at the Alpha Labs of Goertek Inc., where he has worked on XR testing systems and spatial camera projects. His engineering interests span artificial intelligence, data analysis, and the translation of research-grade algorithms into deployable hardware and software.
\end{minipage}

\vspace{1.2em}

\noindent
\begin{minipage}[t]{0.28\linewidth}
  \vspace{0pt}
  \includegraphics[width=\linewidth]{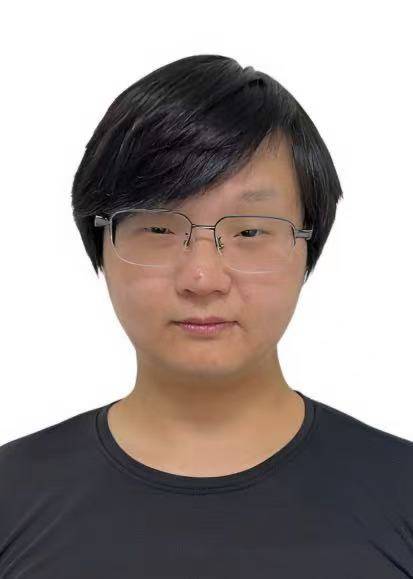}
\end{minipage}\hfill
\begin{minipage}[t]{0.68\linewidth}
  \vspace{0pt}
  \textbf{Ling Zhou} is a Senior Algorithm Engineer at the VRMR center of Goertek Alpha Lab. He is responsible for the research and development of computer vision, image processing, and computer graphics algorithms.
\end{minipage}

\vspace{1.2em}

\end{document}